\documentclass[11pt]{article}

\usepackage{amsmath,amsbsy,amsfonts,amssymb,amsthm,dsfont,fullpage,units}
\usepackage{graphicx,psfrag,epsfig,epsf}
\usepackage{algorithm}
\usepackage{algorithmic,mathtools}
\usepackage{color,cases}
\usepackage{tikz}
\usepackage{enumitem}
\usetikzlibrary{calc,shapes}
\usepackage[round]{natbib}
\usepackage{hyperref}
\usepackage{sublabel}
\usepackage{appendix}
\usepackage{caption,subcaption}

 \newcommand{\IGNORE}[1]{}

\def\nn{\nonumber}

\newcommand\E{\mathbb{E}}
\newcommand\R{\mathbb{R}}

\DeclareMathOperator{\Stopping}{S}

\def\wstar{w}
\def\astar{a}
\def\bstar{b}
\def\cstar{c}

\newcommand\eps{\varepsilon}
\newcommand\veps{\varepsilon}

\newcommand\tl{\tilde}

\newcommand\poly{\operatorname{poly}}

\def\tl{\tilde}

\newcommand\inner[1]{\ensuremath{\langle #1 \rangle}}

\DeclareMathOperator{\polylog}{polylog}

\DeclareMathOperator{\dist}{dist}

\def\tha{{\mbox{\tiny th}}}

 \def\0{{\bf 0}}

\def\viz{{viz.,\ \/}}

\def\nn{\nonumber}

\def\qed{\hfill\hbox{${\vcenter{\vbox{
    \hrule height 0.4pt\hbox{\vrule width 0.4pt height 6pt
    \kern5pt\vrule width 0.4pt}\hrule height 0.4pt}}}$}}

\definecolor{myred}{rgb}{0.3,0.0,0.7}
\definecolor{dkg}{rgb}{0.1,0.7,0.2}
\definecolor{dkb}{rgb}{0.0,0.2,0.8}

 \def\ha{\hat{a}}
 \def\hb{\hat{b}}
 \def\hc{\hat{c}}

 \def\hw{\hat{w}}

\def\hT{\hat{T}}

\def\Sc{{\cal S}}

\def\Ebb{{\mathbb E}}

\def\Rbb{{\mathbb R}}

\newcommand{\bprfof}{\begin{proof_of}}
\newcommand{\eprfof}{\end{proof_of}}
\newcommand{\bprf}{\begin{myproof}}
\newcommand{\eprf}{\end{myproof}}
\newcommand{\bp}{\begin{psfrags}}
\newcommand{\ep}{\end{psfrags}}
\newcommand{\bl}{\begin{lemma}}
\newcommand{\el}{\end{lemma}}
\newcommand{\bt}{\begin{theorem}}
\newcommand{\et}{\end{theorem}}
\newcommand{\bc}{\begin{center}}
\newcommand{\ec}{\end{center}}
\newcommand{\bi}{\begin{itemize}}
\newcommand{\ei}{\end{itemize}}
\newcommand{\ben}{\begin{enumerate}}
\newcommand{\een}{\end{enumerate}}
\newcommand{\bd}{\begin{definition}}
\newcommand{\ed}{\end{definition}}
\def\beq{\begin{equation}}
\def\eeq{\end{equation}\noindent}
\def\beqn{\begin{eqnarray}}
\def\eeqn{\end{eqnarray} \noindent}
\def\beqnn{  \begin{eqnarray*}}
\def\eeqnn{\end{eqnarray*}  \noindent}
\def\bcase{  \begin{numcases}}
\def\ecase{\end{numcases}   \noindent}
\def\bsbcase{  \begin{subnumcases}}
\def\esbcase{\end{subnumcases}   \noindent}

\newtheorem{theorem}{Theorem}

\newtheorem{lemma}{Lemma}

\newtheorem{claim}{Claim}

\newtheorem{definition}{Definition}

\theoremstyle{remark} 
\newtheorem{remark}{Remark}

\newenvironment{myproof}{\noindent{\bf Proof:} \hspace*{1em}}{
    \hspace*{\fill} $\Box$ }
\newenvironment{proof_of}[1]{\noindent {\bf Proof of #1: }}{\hspace*{\fill} $\Box$ }

\newcommand{\matplottc}[1]{               
        \unitlength .45truein
        \begin{center}
        \includegraphics{#1.ps}
        \end{picture}
        \end{center}
}

\def\psfancypar#1#2{\begingroup\def\par{\endgraf\endgroup\lineskiplimit=0pt}
               \setbox2=\hbox{\large\sc #2}
               \newdimen\tmpht \tmpht \ht2 \advance\tmpht by \baselineskip
               \font\hhuge=Times-Bold at \tmpht
               \setbox1=\hbox{{\hhuge #1}}
               \count7=\tmpht \count8=\ht1
               \divide\count8 by 1000 \divide\count7 by \count8
               \tmpht=.001\tmpht\multiply\tmpht by \count7
               \font\hhuge=Times-Bold at \tmpht
               \setbox1=\hbox{{\hhuge #1}}
               \noindent
                \hangindent1.05\wd1
               \hangafter=-2 {\hskip-\hangindent
               \lower1\ht1\hbox{\raise1.0\ht2\copy1}%
                \kern-0\wd1}\copy2\lineskiplimit=-1000pt}

\def\Kout{\setbox1=\hbox{\Huge\bf K}\hbox to
1.05\wd1{\hspace{.05\wd1}
}}
\def\Sout{\setbox1=\hbox{\Huge\bf S}\hbox to 1.05\wd1{\hspace{.05\wd1}
}}

\author{Anima Anandkumar\footnote{University of California, Irvine. Email: a.anandkumar@uci.edu} \and Rong Ge\footnote{Microsoft Research, New England. Email: rongge@microsoft.com} \and Majid Janzamin\footnote{University of California, Irvine. Email: mjanzami@uci.edu}}

\title{Sample Complexity Analysis for Learning Overcomplete \\ Latent Variable Models through Tensor Methods}

\begin{document}
\maketitle

\begin{abstract}
We provide guarantees for learning latent variable models  emphasizing on the overcomplete regime, where the dimensionality of the latent space can exceed  the observed dimensionality. In particular, we consider multiview mixtures, spherical Gaussian mixtures, ICA, and sparse coding models. We provide tight concentration bounds for empirical moments  through novel covering arguments. We analyze parameter recovery through a simple tensor power update algorithm.
In the semi-supervised setting, we exploit the label or prior information to get a rough estimate of the model parameters, and then refine it using the tensor method on unlabeled samples. We establish that learning is possible   when the number of components scales as $k=o(d^{p/2})$, where $d$ is the observed dimension, and $p$ is the order of the observed moment employed in the tensor method. Our concentration bound analysis also leads to minimax sample complexity for semi-supervised learning of spherical Gaussian mixtures.
In the unsupervised setting, we use a simple initialization algorithm based on SVD of the tensor slices, and provide guarantees under the stricter condition that $k\le \beta d$ (where constant $\beta$ can be larger than $1$), where the tensor method recovers the components under a polynomial running time (and exponential in $\beta$). Our analysis establishes that a wide range of overcomplete latent variable models can be learned efficiently with low computational and sample complexity through tensor decomposition methods.
\end{abstract}



\section{Introduction}

It is imperative to incorporate latent variables in any modeling framework. Latent variables can  capture the effect of hidden causes which are not directly observed. Learning these hidden factors is central to many applications, e.g., identifying the latent diseases through observed symptoms, identifying the latent communities through observed social ties, and so on. Moreover, latent variable models (LVMs) can   provide an efficient representation of the observed data, and learning these representations can lead to improved performance on various tasks such as classification. The recent performance gains in  domains such as speech and computer vision can be largely attributed to efficient representation learning~\citep{bengio2012unsupervised}.
Moreover, it has been shown that learning overcomplete representations is crucial to achieving these impressive gains~\citep{coates2011analysis}.

In an overcomplete representation, the dimensionality of the latent space exceeds the observed dimensionality. Overcomplete representations are known to be more robust to noise, and can provide greater flexibility in modeling~\citep{lewicki2000learning}.
Although overcomplete representations have led to huge performance gains in practice, theoretical guarantees for learning are mostly lacking. In many domains, we face the challenging task of   unsupervised or semi-supervised learning, since it is expensive to obtain labeled samples and we typically have access to a large number of   unlabeled samples, e.g.~\citep{coates2011analysis,le2011ica}. Therefore, it is imperative to  develop novel guaranteed methods for efficient unsupervised/semi-supervised learning of overcomplete models.

In this paper, we bridge the gap  between theory and practice, and establish that a wide range of overcomplete LVMs can be learned efficiently through simple spectral learning techniques. We perform spectral decomposition of the higher order moment tensors (estimated using unlabeled samples) to obtain the model parameters.  A recent line of work has shown that tensor decompositions can be employed for unsupervised learning of a wide range of LVMs, e.g., independent components~\citep{de2007fourth}, topic models, Gaussian mixtures, hidden Markov models~\citep{AnandkumarEtal:tensor12}, network community models~\citep{AnandkumarEtal:community12},  and so on.
It involves decomposition of a multivariate moment tensor, and  is guaranteed to provide a consistent estimate of the model parameters.  The sample and computational requirements are only a low order  polynomial in the latent dimensionality for the tensor method~\citep{AnandkumarEtal:tensor12,SongEtal:NonparametricTensorDecomp}.
However, a major drawback behind these works is that they  mostly consider the undercomplete setting, where the latent dimensionality cannot exceed the observed dimensionality.


In this work, we establish guarantees for tensor decomposition in learning overcomplete LVMs, such as multiview mixtures, independent component analysis, Gaussian mixtures and sparse coding models. Note that learning general overcomplete models is ill-posed since the latent dimensionality exceeds the observed dimensionality. We impose a natural incoherence condition  on the components, which can be viewed as a {\em soft orthogonality} constraint, which limits the redundancy among the components. We establish that this constraint not only makes learning well-posed but also enables efficient learning through tensor methods.  Incoherence constraints are natural  in the  overcomplete regime, and have been  considered before, e.g., in compressed sensing~\citep{donoho2006compressed}, independent component analysis~\citep{le2011ica}, and sparse coding~\citep{Arora2013,AgarwalEtal:SparseCoding2013}.



\subsection{Summary of results}
In this paper, we provide semi-supervised and unsupervised learning guarantees for LVMs such as multiview   mixtures, Independent Component Analysis (ICA), Gaussian mixtures and sparse coding models.
For the learning algorithm, we exploit the tensor decomposition algorithm in~\citep{AltTensorDecomp2014}, which   performs alternating asymmetric power updates on the input tensor modes (or performs symmetric power updates if the input tensor is symmetric).
Under the semi-supervised setting, we establish that highly overcomplete models can be learned efficiently through tensor decomposition methods. The moment tensors are constructed using unlabeled samples, and the labeled samples are used to provide a rough initialization to the tensor decomposition algorithm. In the unsupervised setting, we propose a simple initialization strategy for the tensor method, and require  stricter conditions on the extent of overcompleteness for guaranteed learning. In addition, we provide tight concentration bounds on the empirical tensors  through novel covering arguments, which imply efficient sample complexity bounds for learning using the tensor method.


We now summarize the results for learning multiview mixtures model with incoherent components\footnote{We use the term incoherence to say that the deterministic condition in the appendix of~\cite{AltTensorDecomp2014} is satisfied which basically imposes soft-orthogonality constraints on the components. It is also shown that this condition is satisfied whp when the components are uniformly i.i.d. drawn from unit sphere.}.
Let $k$ be the number of hidden components, and $d$ be the observed dimensionality. In the semi-supervised setting, we prove guaranteed learning when $k=o(d^{p/2})$, where   $p$ is the order of observed moment employed for tensor decomposition.  We prove that in the ``low'' noise regime (where the   norm of noise is of the same order as that of the component means), having an extremely small  number of labeled samples for each label  is sufficient (scaling as $\polylog(d,k)$ independent of the final precision). This is far less than the number of unlabeled samples required.
Note that in most applications, labeled samples are expensive/hard to obtain, while many more unlabeled samples are easily available, e.g., see~\cite{le2011ica,CoatesNgLee11}.
Furthermore, we show that the sample complexity bounds for unlabeled samples is $\tl{\Omega}(k)$. Note that this   is the {\em minimax} bound  up to $\polylog$ factors.

We also provide {\em unsupervised} learning guarantees when no label is available.
Here, the initialization is obtained  by performing a rank-$1$ SVD on the random slices of the moment tensor. This imposes additional conditions on rank and sample complexity. We prove  that when $k\le \beta d$ (for arbitrary constant $\beta$ which can be larger than 1), the model parameters can be learned using a polynomial number of initializations (which depends on $\beta$ and scales as $k^{\beta^2}$) and sample complexity scales as $\tl{\Omega}(kd)$, which is efficient.


We also provide  semi-supervised and unsupervised learning guarantees for ICA model. By semi-supervised setting in ICA, we mean some prior information is available which provides good initializations (with a constant $\ell_2$ error on the columns) for the tensor decomposition algorithm.
In the semi-supervised setting, we show that when the number of components scales as $k = \Theta(d^2)/\polylog(d)$, the ICA model can be efficiently learned from fourth order moments with $n \geq \tl{\Omega}(k^{2.5})$ number of unlabeled samples.  In the unsupervised setting, we show that when $k=\Theta(d)$, the ICA model can be learned with $n \geq \tl{\Omega}(k^3)$ in time $k^{\Omega(k^2/d^2)}$.



We also provide learning results for the sparse coding model, when the coefficients are independently drawn from a Bernoulli-Gaussian distribution. 
Note that this corresponds to a sparse ICA model since the hidden coefficients are independent.  Let $s$ be the expected sparsity level of the hidden variables. In the semi-supervised setting (where prior information gives good initialization), we require $\tilde{\Omega}(\max\{sk,s^2k^2/d^3\})$ number of unlabeled samples for learning as long as $k=o(d^2)$. Note that in the special case when $s$ is a constant, the sample complexity is akin to learning multiview models, where $s=1$; and when $s=\Theta(k)$, it is akin to learning the ``dense'' ICA model, where $s=k$. Thus, the sparse coding model bridges the range of models  between multiview mixtures  and ICA. Furthermore, we also extend the learning results to dependent sparsity setting, but with worse performance guarantees.

Although we prove strong theoretical guarantees for learning overcomplete models, there are two main  caveats for our approach. We recover the model parameters with an {\em approximation } error, which decays with the dimension $d$. Concretely, for the $p^{\tha}$ order tensor, the approximation error is
$\tilde{O} \bigl( \sqrt{k/d^{p-1}} \bigr)$, which decays since $k = o(d^{p/2})$. This is because  the actual mixture components are not the stationary points of the tensor algorithm updates (even in the noiseless setting)
since the components are not  strictly orthogonal. This bias can be presumably removed by performing joint updates (e.g alternating least squares) where the objective is to fit the learnt vectors to the input tensor and we leave this for future study.  Second, the setting is not suited for topic models, where there is a non-negativity constraint on the topic-word matrix. Here, incoherence can only be enforced through sparsity, and since our method does not exploit sparsity, we believe that other formulations may be better suited for learning in this setting.

\paragraph{Overview of techniques:} We establish tight concentration bounds for empirical tensors when the samples are drawn from  multiview linear mixtures, Gaussian mixtures, ICA or sparse coding models.  The concentration bound involves bounding the spectral norm of the error tensor, and this relies on the construction of  {\em $\eps$-nets} to cover all vectors (on the sphere). A naive $\eps$-net argument is however too loose since it results in a large number of vectors without a ``fine-grained'' distinction between them.  A more refined notion is to employ an {\em entropy-concentration} trade-off, as proposed in~\cite{RudelsonVershynin2009}, where the   vectors in the $\eps$-net are classified into sparse and dense vectors, and to analyze them separately. The sparse vectors can result in   large correlations, but  the number of such vectors is small, while the dense vectors have small correlations, although their number is larger. In our setting, however, this classification is still not enough, and we need a more refined analysis. We group the data samples into ``buckets'' based on their correlation with a given vector, and  bound each ``bucket''   separately. We impose additional conditions on the factor and noise matrices to bound the size of the buckets.

For the multiview linear mixtures, we impose a restricted isometry property  (RIP) on the noise matrices and a  bounded   $2 \to 3$ norm condition on the factor matrices (which is weaker than RIP). For Gaussian mixtures, the RIP property on noise is satisfied, and we only require a condition of bounded $2\to 3$ norm on the matrix of component mean vectors. These constraints allow us to bound the size of the ``buckets'', where each bucket corresponds to noise or factor vectors   with a certain level of correlation with a fixed vector. Intuitively, the number of samples having a high correlation with a fixed vector (i.e. size of a ``bucket'') cannot be too large due to RIP/bounded 2-to-3 norm constraints. We apply Bernstein's bound on each of these buckets separately and combine them to obtain the final bound.  Our construction has only a logarithmic number of buckets (since we vary correlation levels geometrically), and therefore the overall concentration bound only has additional logarithmic factors when we combine the results.

For the ICA model, the conditions and analysis are somewhat different.  This is because all the hidden sources ``mix'' together in each sample, in contrast to the mixture model, where each sample is generated from only one component. Establishing concentration bounds involves two steps, \viz first having a bound on the fourth order empirical moment of the hidden sources, assuming they are sub-Gaussian and kurtotic,\footnote{Note that while the kurtotis (4th order cumulant) of a Gaussian random variable is zero, the kurtotis of sub-Gaussian random variables is in general nonzero. In addition, note that this analysis can be also extended to sub-exponential random variables.} and then converting the bound to  the observed space. This involves a   spectral norm bound on the linear map between the hidden sources and the observations.

We then consider the sparse coding model, where the hidden variables are assumed to be sparsely activated. In the special case, when the hidden variables are independent, this corresponds to a sparse ICA model. We derive the concentration bound   for Bernoulli-Gaussian variables, assuming that the dictionary has the RIP property (e.g., Gaussian matrix). In this case, we establish that the concentration bound depends only on the sparsity level, and not on the total number of dictionary elements. Here, we partition the vectors into ``buckets'' based on their correlation with the dictionary elements and the RIP property allows us to bound the size of buckets, as before in the case of multiview mixtures. In addition, we exploit the sparsity of elements to obtain a tighter bound for the sparse coding setting.

Thus, we obtain tight concentration bounds for empirical tensors for multiview and Gaussian mixtures,  ICA and sparse coding models.   The conditions on noise (RIP) and factor matrices (bounded $2$-to-$3$ norm) are fairly benign and natural to impose.  Our novel bucketing arguments could be applicable in other settings involving matrix and tensor concentration bounds.

We then employ the concentration bounds in conjunction with   the alternating rank-$1$  updates algorithm to obtain learning guarantees for the above models. In our recent work~\citep{AltTensorDecomp2014}, we establish  local and global convergence guarantees for this algorithm  when the components are incoherent.  We  combine these guarantees with the concentration bounds to establish that a wide range of latent variable models can be learned with low computational and sample complexities.


\subsection{Related work}

\paragraph{Tensor decomposition for learning undercomplete models: }Several latent variable models can be learned through tensor decomposition including independent component analysis~\citep{de2007fourth}, topic models, Gaussian mixtures, hidden Markov models~\citep{AnandkumarEtal:tensor12} and network community models~\citep{AnandkumarEtal:community12}.
In the undercomplete setting, \citet{AnandkumarEtal:tensor12} analyze robust tensor power iteration for learning LVMs,  and \citet{SongEtal:NonparametricTensorDecomp} extend analysis to the nonparametric setting. These works require the tensor factors to have full column rank, which rules out overcomplete models.
Moreover, they require whitening the input data, and hence the sample complexity depends on the condition number of the factor matrices.
For instance, when $k = d$, for random factor matrices, the previous tensor approaches in~\citet{SongEtal:NonparametricTensorDecomp,AnandkumarEtal:lda12} have a sample complexity of  $\tl{\Omega} (k^{6.5})$, while our result provides improved sample complexity $\tl{\Omega}(k^2)$ assuming incoherent components.



\paragraph{Learning overcomplete models: }In general, learning overcomplete models is challenging, and they may not even be identifiable. The FOOBI procedure by~\citet{de2007fourth} shows that a polynomial-time procedure can recover the components of ICA model (with {\em generic} factors) when $k =O(d^2)$, where the moment is fourth order. However, the procedure does not work for third-order overcomplete tensors.
For the fifth order tensor, \citet{fourierpca,bhaskara2013smoothed} perform simultaneous diagonalization on the matricized versions of random slices of the tensor and provide careful perturbation analysis. But, this  procedure cannot handle the same level of overcompleteness as FOOBI, since an additional dimension is required for obtaining two (or more) fourth order tensor slices.
In addition, \citet{fourierpca} provide stronger results for ICA, where the tensor slices can be obtained in the Fourier domain. Given $4$th order tensor, they need $\poly(k^4)$ number of unlabeled samples for learning ICA (where the poly factor is not explicitly characterized), while we only need $\tl{\Omega}(k^{2.5})$ (when $k = \Theta(d^2)/\polylog(d)$).
\citet{GMMICA2013} convert the problem of learning Gaussian mixtures to an ICA problem and exploit the Fourier PCA method in \citet{fourierpca}. More precisely, for a Gaussian mixtures model with known identical covariance matrices,   when the number of components $k =\poly(d)$, the model can be learned in polynomial time (as long as a certain non-degeneracy condition is satisfied).



~\citet{Arora2013,AgarwalEtal:SparseCoding2013,barak2014} provide guarantees for the sparse coding model (also known as dictionary learning  problem).
\citet{Arora2013,AgarwalEtal:SparseCoding2013} provide clustering based approaches for approximately learning incoherent dictionaries and then refining them through alternating minimization to obtain exact recovery of both the dictionary and the coefficients. They can handle sparsity level up to $O(\sqrt{d})$ (per sample) and the size of the dictionary $k$ can be arbitrary.
\citet{barak2014} consider tensor decomposition and dictionary learning using sum-of-squares (SOS) method. In contrast to simple iterative updates considered here, SOS involves solving semi-definite programs.
They provide guaranteed recovery by a polynomial time complexity $k^{O(1/\delta)}$ for some $0<\delta<1$, when the size of the dictionary   $k = \Theta(d)$, and  the sparsity level is $k^{1-\delta}$.
They also provide guarantees for higher sparsity levels up to (a small enough) constant  fraction of $k$, but the computational complexity of the algorithm becomes  quasi-polynomial: $k^{O(\log k)}$.
They can also handle  higher level of overcompleteness at the expense of reduced  sparsity level.
They do not require any incoherence conditions on the factor matrices and they can handle the signal to  noise ratio  being a constant. Thus, their work has strong guarantees, but at the expense of running a complicated algorithm. In contrast, we consider a simple alternating rank-$1$ updates algorithm, but require more stringent conditions on the model.


There are other recent works which can learn overcomplete models, but under different settings than the one considered in this paper. \citet{AnandkumarEtal:NIPS13} learn overcomplete sparse topic models, and provide guarantees for {\em Tucker} tensor decomposition under sparsity constraints. Specifically, the model is identifiable using $(2n)^{\tha}$ order moments when the latent dimension  $k=O(d^n)$ and   the sparsity level of the factor matrix is   $O(d^{1/n})$, where $d$ is the observed dimension.
The Tucker decomposition is more general than the CP decomposition considered here, and the techniques in~\citep{AnandkumarEtal:NIPS13} differ significantly from the ones considered here, since they incorporate sparsity, while we incorporate incoherence here.

\paragraph{Concentration Bounds: }We obtain tight concentration bounds for empirical tensors in this paper. In contrast, applying matrix concentration bounds, e.g.~\citep{RandomMatrices:Tropp}, leads to strictly worse bounds since they require matricizations of the tensor.
\citet{LatalaBound} provides an upper bound on the moments of the Gaussian chaos, but they are limited to independent Gaussian distributions (and can be extended to other cases such as Rademacher distribution). The principle of entropy-concentration trade-off~\citep{RudelsonVershynin2009}, employed in this paper, have been used in other contexts. For instance,~\citet{TenSparsification} provide a spectral norm bound for random tensors. They first apply a symmetrization argument which reduces the problem to bounding the spectral norm of a random Gaussian tensor and then employ entropy-concentration trade-off to bound its spectral norm. They also exploit the bounds on the Lipschitz functions of Gaussian random variables. While~\citet{TenSparsification} employ a rough classification of vectors (to be covered) into dense and sparse vectors, we require a finer classification of vectors into different ``buckets'' (based on their inner products with given vectors) to obtain the tight concentration bounds in this paper. Moreover, we do not impose Gaussian assumption in this paper, and instead require more general conditions such as RIP or bounded $2$-to-$3$ norms.

\subsection{Notations and tensor preliminaries}
Define $[n] := \{1,2,\dotsc,n\}$.
Let $\|u\|_p$ denote the $\ell_p$ norm of vector $u$, and the induced $q \rightarrow p$ norm of matrix $A$ is defined as
$$
\|A\|_{q \to p} := \sup_{\|u\|_q = 1} \|A u\|_p.
$$

Notice that while the standard asymptotic notation is to write $f(d) = O(g(d))$ and $g(d) = \Omega(f(d))$, we sometimes use $f(d) \leq O(g(d))$ and $g(d) \geq \Omega(f(d))$ for additional clarity.
We also use the asymptotic notation $f(d) = \tl{O}(g(d))$ if and only if $f(d) \leq \alpha g(d)$ for all $d \geq d_0$, for some $d_0 >0$ and $\alpha = \polylog(d)$, i.e., $\tilde{O}$ hides $\polylog$ factors. Similarly, we say $f(d) = \tl{\Omega}(g(d))$ if and only if $f(d) \geq \alpha g(d)$ for all $d \geq d_0$, for some $d_0 >0$ and $\alpha = \polylog(d)$.

\subsubsection*{Tensor preliminaries}
A real \emph{$p$-th order tensor} $T \in \bigotimes_{i=1}^p \R^{d_i}$ is a member of the outer product of Euclidean spaces $\R^{d_i}$, $i \in [p]$.
For convenience, we restrict to the case where $d_1 = d_2 = \dotsb = d_p = d$, and simply write $T \in \bigotimes^p \R^d$.
As is the case for vectors (where $p=1$) and matrices (where $p=2$), we may
identify a $p$-th order tensor with the $p$-way array of real numbers $[
T_{i_1,i_2,\dotsc,i_p} \colon i_1,i_2,\dotsc,i_p \in [d] ]$, where
$T_{i_1,i_2,\dotsc,i_p}$ is the $(i_1,i_2,\dotsc,i_p)$-th coordinate of $T$
with respect to a canonical basis. For convenience, we limit to third order tensors $(p=3)$ for the rest of this section, while the results for higher order tensors are similar.

The different dimensions of the tensor are referred to as {\em modes}. For instance, for a matrix, the first mode refers to columns and the second mode refers to rows.
In addition,￼ {\em fibers} are higher order analogues of matrix rows and columns. A fiber is obtained by fixing all but one of the indices of the tensor (and is arranged as a column vector). For instance, for a matrix, its mode-$1$ fiber is any matrix column while a mode-$2$ fiber is any row. For a
third order tensor $T\in \R^{d \times d \times d}$, the mode-$1$ fiber is given by $T(:, j, l)$, mode-$2$ by $T(i, :, l)$ and mode-$3$ by $T(i, j, :)$.
Similarly, {\em slices} are obtained by fixing all but two of the indices of the tensor. For example, for the third order tensor $T$, the slices along $3$rd mode are given by $T(:, :, l)$.

We view a tensor $T \in \Rbb^{d \times d \times d}$ as a multilinear form. Consider matrices $M_r \in \R^{d\times d_r}, r \in \{1,2,3\}$. Then tensor $T(M_1,M_2,M_3) \in \R^{d_1}\otimes \R^{d_2}\otimes \R^{d_3}$ is defined as
\begin{align} \label{eqn:multilinear form def}
T(M_1,M_2,M_3)_{i_1,i_2,i_3} := \sum_{j_1, j_2,j_3\in[d]} T_{j_1,j_2,j_3} \cdot M_1(j_1, i_1) \cdot M_2(j_2, i_2) \cdot M_3(j_3, i_3).
\end{align}
In particular, for vectors $u,v,w \in \R^d$, we have\,\footnote{Compare with the matrix case where for $M \in \R^{d \times d}$, we have $ M(I,u) = Mu := \sum_{j \in [d]} u_j M(:,j) \in \R^d$.}
\begin{equation} \label{eqn:rank-1 update}
 T(I,v,w) = \sum_{j,l \in [d]} v_j w_l T(:,j,l) \ \in \R^d,
\end{equation}
which is a multilinear combination of the tensor mode-$1$ fibers.
Similarly $T(u,v,w) \in \R$ is a multilinear combination of the tensor entries,  and $T(I, I, w) \in \R^{d \times d}$ is a linear combination of the tensor slices.

A $3$rd order tensor $T \in \Rbb^{d \times d \times d}$ is said to be rank-$1$ if it can be written in the form
\begin{align} \label{eqn:rank-1 tensor}
T= w \cdot a \otimes b\otimes c \Leftrightarrow T(i,j,l) = w \cdot a(i) \cdot b(j) \cdot c(l),
\end{align}
where notation $\otimes$  represents the {\em outer product} and $a \in \Rbb^d$, $b \in \Rbb^d$, $c \in \Rbb^d$ are unit vectors (without loss of generality).
A tensor $T  \in \Rbb^{d \times d \times d}$ is said to have a CP rank $k\geq 1$ if it can be written as the sum of $k$ rank-$1$ tensors
\begin{equation}\label{eqn:tensordecomp}
T = \sum_{i\in [k]} w_i a_i \otimes b_i \otimes c_i, \quad w_i \in \Rbb, \ a_i,b_i,c_i \in \Rbb^d.
\end{equation}
This decomposition is closely related to the multilinear form. In particular, for vectors $\ha,\hb,\hc \in \Rbb^d$, we have
$$T(\ha,\hb,\hc) = \sum_{i\in [k]} w_i \langle a_i, \ha\rangle\langle b_i, \hb\rangle\langle c_i,\hc\rangle.$$
Consider the decomposition in equation~\eqref{eqn:tensordecomp},
denote matrix $A:=[a_1 \ a_2 \ \dotsb \ a_k] \in \R^{d \times k}$, and similarly $B$ and $C$. Without loss of generality, we assume that the matrices have normalized columns (in $2$-norm), since we can always rescale them, and adjust the weights $w_i$ appropriately.

For vector $v \in \R^d$, we define $$v^{\otimes p} := v \otimes v \otimes
\dotsb \otimes v \in \bigotimes^p \R^d$$ as its $p$-th tensor power.

Throughout, $\|v\| := (\sum_i v_i^2)^{1/2}$ denotes the Euclidean or $\ell_2$ norm
of a vector $v$, and $\|M\|$ denotes the spectral (operator) norm of a matrix $M$.
Furthermore, $\|T\|$ and $\|T\|_F$ denote the spectral (operator) norm and the Frobenius norm of a tensor, respectively. In particular, for a $3$rd order tensor, we have
$$
\|T\| := \sup_{\|u\| = \|v\| = \|w\| = 1} |T(u,v,w)|, \quad \|T\|_F := \sqrt{\sum_{i,j,l \in [d]} T_{i,j,l}^2}.
$$

\section{Tensor Decomposition for Learning Latent Variable Models}  \label{sec:LVMs}

In this section, we discuss that the problem of learning several latent variable models reduces to the tensor decomposition problem. We show that the observed moment of the latent variable models can be written in a CP tensor decomposition form when appropriate modifications are performed. This is done for multiview linear mixtures model, spherical Gaussian mixtures and ICA (Independent Component Analysis). 
For a more detailed discussion on the connection between observed moments of LVMs and tensor decomposition, see Section 3 in~\citet{AnandkumarEtal:tensor12}.

Therefore, an efficient tensor decomposition method leads to efficient learning procedure for a wide range of latent variable models. In Section~\ref{sec:algorithm}, we provide the tensor decomposition algorithm introduced in \citet{AltTensorDecomp2014}, and exploit it for learning latent variable models providing sample complexity results in the subsequent sections. Note that the sample complexity guarantees are argued through tensor concentration bounds proposed in Section~\ref{sec:Tensor Concent.}.

\subsection{Multiview linear mixtures model} \label{sec:multiview}

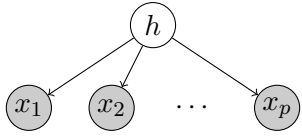
\begin{figure}
\begin{center}
\begin{tikzpicture}
  [
    scale=1.1,
    observed/.style={circle,minimum size=0.6cm,inner
sep=0mm,draw=black,fill=black!20},
    hidden/.style={circle,minimum size=0.6cm,inner sep=0mm,draw=black},
  ]
  \node [hidden,name=h] at ($(0,0)$) {$h$};
  \node [observed,name=x1] at ($(-1.5,-1)$) {$x_1$};
  \node [observed,name=x2] at ($(-0.5,-1)$) {$x_2$};
  \node [observed,name=xl] at ($(1.5,-1)$) {$x_p$};
  \node at ($(0.5,-1)$) {$\dotsb$};
  \draw [->] (h) to (x1);
  \draw [->] (h) to (x2);
  \draw [->] (h) to (xl);
\end{tikzpicture}
\end{center}
\caption{\small Multi-view mixtures model.}
\label{fig:Multiview}
\end{figure}

Consider a multiview linear mixtures model as in Figure \ref{fig:Multiview} with $k$ components and $p \geq 3$ views. Throughout the paper, we assume $p=3$ for simplicity, while the results can be also extended to higher-order. Suppose that hidden variable $h \in [k]$ is a discrete categorical random variable with $\Pr [h = j] = w_j, j \in [k]$.
The variables (views) $x_l \in \Rbb^d$ are conditionally independent given the $k$-categorical latent variable $h \in [k]$, and the conditional means are
\begin{align*}
\Ebb[x_1|h] = a_h, \quad \Ebb[x_2|h] = b_h, \quad \Ebb[x_3|h] = c_h,
\end{align*}
where $A := [a_1 \ a_2 \ \dotsb \ a_k ] \in \R^{d \times k}$ denotes the {\em factor matrix} and $B, C$ are similarly defined. 
The goal of the learning problem is to recover the parameters of the model (factor matrices) $A$, $B$, and $C$ given observations.

For this model, the third order observed moment has the form (See~\citealt{AnandkumarEtal:tensor12})
\begin{equation}\label{eqn:tensordecompMixture}
\Ebb [x_1 \otimes x_2 \otimes x_3] = \sum_{j \in [k]} w_j a_j \otimes b_j \otimes c_j.
\end{equation}
The decomposition in \eqref{eqn:tensordecompMixture} is referred to as the CP decomposition~\citep{carroll1970analysis}, and $k$ denotes the CP tensor rank.
Hence, given third order observed moment, the unsupervised learning problem (recovering factor matrices $A$, $B$, and $C$) reduces to computing a tensor decomposition as in \eqref{eqn:tensordecompMixture}. 

In addition, suppose that given hidden state $h$, the observed variables $x_l \in \R^d$ have conditional distributions as
\begin{align*}
x_1|h\sim a_h + \zeta \sqrt{d} \cdot \veps_A, \quad x_2|h\sim b_h + \zeta \sqrt{d} \cdot \veps_B, \quad x_3|h\sim c_h + \zeta \sqrt{d} \cdot \veps_C,
\end{align*}
where $\veps_A,\veps_B,\veps_C \in \Rbb^d$ are independent random vectors with zero mean and covariance $\frac{1}{d} I_d$, and $\zeta^2$ is a scalar denoting the variance of each entry. We also assume that noise vectors $\veps_A,\veps_B,\veps_C$ are independent of hidden vector $h$. In addition, let
all the vectors $a_h,b_h,c_h, h \in [k],$ have unit $\ell_2$ norm. Furthermore, since $w_j$'s are the mixture probabilities, for simplicity we consider $w_j = \Theta(1/k), j\in [k]$.
We call this model $\mathcal{S}$.

When $\zeta^2 = \Theta(1/d)$, the norm of the noise is roughly the same as the norm of the components. We call this the {\em low noise regime}. When $\zeta^2 = \Theta(1)$, the norm of noise in {\em every dimension} is roughly the same as the norm of the components. We call this the {\em high noise regime}.

\subsection{Spherical Gaussian mixtures} \label{sec:SphericalGaussMix}
Consider a mixture of $k$ different Gaussian distributions with spherical covariances. Let $w_j, j \in [k]$ denote the proportion for choosing each mixture. For each Gaussian component $j \in [k]$, $a_j \in  \Rbb^d$ is the mean, and $\zeta_i^2 I$ is the spherical covariance.
For simplicity, we restrict to the case where all the components have the same spherical variance, i.e., $\zeta_1^2 = \zeta_2^2 = \dotsb = \zeta_k^2 = \zeta^2$. The generalization is discussed in \citet{SphericalGaussian2012}. In addition, in order to generalize the learning result to the overcomplete setting, we assume that variance parameter $\zeta^2$ is known (see Remark \ref{remark:variance SGM} for more discussions).
The following lemma shows that the problem of estimating parameters of this mixture model can be formulated as a tensor decomposition problem. This is a special case of Theorem 1 in \citet{SphericalGaussian2012} where we assume the variance parameter is known.

\begin{lemma}[\citealt{SphericalGaussian2012}]
If
\begin{align}
M_3 & := \Ebb[x \otimes x \otimes x] - \zeta^2 \sum_{i \in [d]} \left( \Ebb[x] \otimes e_i \otimes e_i + e_i \otimes \Ebb[x] \otimes e_i + e_i \otimes e_i \otimes \Ebb[x] \right), \label{eqn:spherical Gaussian modified moment}
\end{align}
then
\begin{align*}
M_3 &= \sum_{j \in [k]} w_j a_j \otimes a_j \otimes a_j.
\end{align*}
\end{lemma}

In order to provide the learning guarantee, we define the following empirical estimates. Let $\mathcal{\widehat{M}}_3$, $\mathcal{\widehat{M}}_2$, and $\mathcal{\widehat{M}}_1$ respectively denote the empirical estimates of the raw moments $\Ebb[x \otimes x \otimes x]$, $\Ebb[x \otimes x]$, and $\Ebb[x]$.
Then, the empirical estimate of the third order modified moment in \eqref{eqn:spherical Gaussian modified moment} is
\begin{align} \label{eqn:spherical Gaussian_empirical modified moment}
\widehat{M}_3 & := \mathcal{\widehat{M}}_3 - \zeta^2 \sum_{i \in [d]} \left( \mathcal{\widehat{M}}_1 \otimes e_i \otimes e_i + e_i \otimes \mathcal{\widehat{M}}_1 \otimes e_i + e_i \otimes e_i \otimes \mathcal{\widehat{M}}_1 \right).
\end{align}

\begin{remark}[Variance parameter estimation] \label{remark:variance SGM}
Notice that we assume variance $\zeta^2$ is known in order to generalize the learning result to the overcomplete setting. Since $\zeta$ is a scalar parameter, it is reasonable to try different values of $\zeta$ till we get a good reconstruction.
On the other hand, in the undercomplete setting, variance $\zeta^2$ can be also estimated as proposed in \cite{SphericalGaussian2012}, where estimate $\hat{\zeta}^2$ is the $k$-th largest eigenvalue of the empirical covariance matrix $\mathcal{\widehat{M}}_2 - \mathcal{\widehat{M}}_1 \mathcal{\widehat{M}}_1^{\,\top}$.
\end{remark}

\subsection{Independent component analysis (ICA)} \label{sec:ICA}
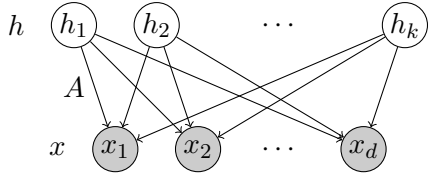
\begin{figure}
\begin{center}
\begin{tikzpicture}
  [
    scale=1.1,
    observed/.style={circle,minimum size=0.6cm,inner
sep=0mm,draw=black,fill=black!20},
    hidden/.style={circle,minimum size=0.6cm,inner sep=0mm,draw=black},
  ]
  \node [hidden,name=h1] at ($(-2,0)$) {$h_1$};
  \node [hidden,name=h2] at ($(-1,0)$) {$h_2$};
  \node [hidden,name=hk] at ($(2,0)$) {$h_k$};
  \node [] at ($(-2.7,0)$) {$h$};
  \node [observed,name=x1] at ($(-1.5,-1.5)$) {$x_1$};
  \node [observed,name=x2] at ($(-0.5,-1.5)$) {$x_2$};
  \node [observed,name=xd] at ($(1.5,-1.5)$) {$x_d$};
  \node [] at ($(-2.2,-1.5)$) {$x$};
  \node [] at ($(-2,-0.75)$) {$A$};
  \node at ($(0.5,-1.5)$) {$\dotsb$};
  \node at ($(0.5,0)$) {$\dotsb$};
  \draw [->] (h1) to (x1);
  \draw [->] (h1) to (x2);
  \draw [->] (h1) to (xd);
  \draw [->] (h2) to (x1);
  \draw [->] (h2) to (x2);
  \draw [->] (h2) to (xd);
  \draw [->] (hk) to (x1);
  \draw [->] (hk) to (x2);
  \draw [->] (hk) to (xd);
\end{tikzpicture}
\end{center}
\caption{\small Graphical representation of ICA model $x=Ah$, where the coordinates of $h$ are independent.}
\label{fig:ICA}
\end{figure}

In the standard ICA model~\citep{Comon94,CardosoComonICA,ICA2000,ComonJuttenICA}, random independent latent signals are linearly mixed and perturbed with noise to generate the observations. Let $h \in \R^k$ be a random latent signal, where its coordinates are independent, $A \in \Rbb^{d \times k}$ be the mixing matrix, and $z \in \Rbb^d$ be the Gaussian noise. In addition, $h$ and $z$ are also independent. Then, the observed random vector is
\begin{align*}
x = Ah + z.
\end{align*}
Figure~\ref{fig:ICA} depicts a graphical representation of the ICA model where the coordinates of $h$ are independent.

The following lemma shows that the problem of estimating parameters of the ICA model can be formulated as a tensor decomposition problem.
\begin{lemma}[\citealt{ComonJuttenICA}]
Define
\begin{align} \label{eqn:ICA_modified moment}
M_4 := \Ebb [x \otimes x \otimes x \otimes x] - T,
\end{align}
where $T \in \R^{d \times d \times d \times d}$ is the fourth order tensor with
\begin{align} \label{eqn:ICA_2nd order term}
T_{i_1,i_2,i_3,i_4} := \Ebb[x_{i_1} x_{i_2}] \Ebb[x_{i_3} x_{i_4}] + \Ebb[x_{i_1} x_{i_3}] \Ebb[x_{i_2} x_{i_4}] + \Ebb[x_{i_1} x_{i_4}] \Ebb[x_{i_2} x_{i_3}], \quad i_1,i_2,i_3,i_4 \in [d].
\end{align}
Let $\kappa_j := \Ebb[h_j^4] - 3 \Ebb^2[h_j^2]$,  $j \in [k]$. Then, we have
\begin{equation} \label{eqn:ICA_tensor form}
M_4 = \sum_{j \in [k]} \kappa_j a_j \otimes a_j \otimes a_j \otimes a_j.
\end{equation}
\end{lemma}
See~\cite{SphericalGaussian2012} for a proof of this theorem in this form.
Let $\widehat{M}_4$ be the empirical estimate of $M_4$ given $n$ samples.

\subsubsection*{Sparse ICA}

We also consider the sparse ICA model, which is the ICA with the additional constraint that the hidden vector $h$ is sparse.

This is related to the dictionary learning or sparse coding model $x=Ah$ where the observations $x \in \R^d$ are sparse combination of dictionary atoms $a_j \in \R^d, j \in [k]$ through sparse vector $h \in \R^k$. If in addition, the coordinates of $h$ are random and independent, the dictionary learning model is the same as the sparse ICA model.
Others have studied the general sparse coding problem which are briefly mentioned in the related works section.


\section{Tensor Concentration Bounds} \label{sec:Tensor Concent.}

In this section, we provide tensor concentration results for the proposed latent variable models. For each LVM, consider the higher-order observed moment (tensor) described in Section~\ref{sec:LVMs}. The tensor concentration result bounds the spectral norm of error between the true moment tensor and its empirical estimate given $n$ samples.

\subsection{Multiview linear mixtures model}

For the multiview linear mixtures model, we provide the tensor concentration result for the $3$rd order observed moment in~\eqref{eqn:tensordecompMixture}. 

Consider the multiview linear mixtures model described in Section~\ref{sec:multiview} denoted as model $\Sc$.  Let $x_1^i, x_2^i, x_3^i, i \in [n]$, denote $n$ samples of views $x_1, x_2, x_3$, respectively.  Since the main focus is on recovering the components, we bound the spectral norm of difference between the empirical tensor estimate
\[\hat{T} := \frac{1}{n} \sum_{i=1}^n x^i_1\otimes x^i_2\otimes x^i_3,\]
and
\[ \tilde{T} := \E \bigl[ x_1 \otimes x_2 \otimes x_3 |h_i, i \in [n] \bigr] = \frac{1}{n} \sum_{i=1}^n (a_{h_i})\otimes (b_{h_i})\otimes (c_{h_i}),\]
where the expectation is conditioned on the choice of hidden states for $n$ samples, and taken over the randomness of noise. Here, $h_i \in [k]$ denotes the hidden state for sample $i \in [n]$.
Notice that tensor $\tilde{T}$ has the same form as true tensor $T$ in \eqref{eqn:tensordecompMixture} where
$$
\tilde{T} = \sum_{j\in [k]} \tilde{w}_j a_j\otimes b_j\otimes c_j.
$$
Here $\tilde{w}_j, j \in [k]$ are the empirical frequencies of different hidden states $h \in [k]$. It is easy to see that if $n \geq \Omega \left( \frac{\log k}{w_{\min}} \right)$, then all the empirical frequencies $\tilde{w}_j$ are within $[w_j/2,2w_j]$. Therefore, tensor decomposition of $\tilde{T}$ has the same eigenvectors and similar eigenvalues as the true expectation (over both the noise and the hidden variables), and hence, it suffices to bound $\|\hat{T} - \tilde{T} \|$ provided as follows.


\begin{theorem}[Tensor concentration bound for multiview linear mixtures model] \label{lem:TenConcentMixtureModel}
Consider $n$ samples $\{(x^i_1, x^i_2,x^i_3), i \in [n]\}$ from the multiview linear mixtures model $\mathcal{S}$ with corresponding hidden states $\{h_i,i \in [n]\}$.
Assume matrices $A^\top$, $B^\top$ and $C^\top$ have $2\to 3$ norm bounded by $O(1)$, and noise matrices $E_A$, $E_B$ and $E_C$ defined in \eqref{eqn:noise matrix} satisfy the RIP condition in \ref{RIP} (see Remark~\ref{remark:RIP} for details on RIP condition). For $\hat{T}$ and $\tilde{T}$ as above, if $n = \poly(d)$, we have with high probability (over the choice of hidden state $h$ and the noise)
$$
\|\hat{T} - \tilde{T} \| \le \tilde{O}\left(\zeta \left(\frac{\sqrt{d}}{n} + \sqrt{w_{\max}\frac{d}{n}}\right) + \zeta^2\left(\frac{d}{n}+\sqrt{w_{\max}\frac{d^{1.5}}{n}}\right)+\zeta^3 \left(\frac{d^{1.5}}{n}+\sqrt{\frac{d}{n}}\right)\right).
$$
\end{theorem}

See the proof in Appendix~\ref{appendix:multiview concent bound}. The main ideas are described later in this section.

The above bound holds for any level of noise, but in each specific regime of noise, one of the terms is dominant and the bound is simplified. We now provide the bound for the high noise $\zeta^2 = \Theta(1)$ and low noise $\zeta^2 = \Theta \left( 1/d \right)$ regimes which were introduced in Section~\ref{sec:multiview}.
In the high noise regime $\zeta^2 = \Theta(1)$, the term $\zeta^3 \sqrt{\frac{d}{n}}$ in Theorem~\ref{lem:TenConcentMixtureModel} is dominant, and in the low noise regime $\zeta^2 = \Theta \left( 1/d \right)$, the term $\zeta \sqrt{w_{\max} \frac{d}{n}}$ in Theorem~\ref{lem:TenConcentMixtureModel} is dominant. This concentration bound is later used in Section~\ref{sec:multiview learning} to provide sample complexity guarantees for learning multiview linear mixtures model.

\begin{remark}[Application of Theorem~\ref{lem:TenConcentMixtureModel} to whitening-based approaches] \label{remark:whitening}
In the undercomplete setting, a guaranteed approach for tensor decomposition is to first orthogonalize the tensor through the {\em whitening} step, and then perform the orthogonal tensor eigen-decomposition through the power method~\citep{AnandkumarEtal:tensor12}. The whitening step leads to dependency to the condition number in the sample complexity result. Applying the proposed tensor concentration bound in Theorem~\ref{lem:TenConcentMixtureModel} to this approach, we get similar dependency to the condition number, but better dependency in the dimension $d$. This improvement comes at the cost of additional bounded $2 \to 3$ norm condition on the factor matrices.

Concretely, following the analysis in~\cite{AnandkumarEtal:tensor12,SongEtal:NonparametricTensorDecomp}, we have the error in recovery (up to permutation) as
\begin{equation}\| \hat{a}_i - a_i \| \leq \frac{32\sqrt{2}\epsilon_{\textnormal{triples}}}{\sigma_{\min}^3 w_{\min}^{1.5}}+ \frac{512 \epsilon^3_{\textnormal{pairs}}}{\sigma^3_{\min} w_{\min}^{1.5}},\label{eqn:errorwhiten}\end{equation}where $\epsilon_{\textnormal{triples}}:= \| \hat{T}- \tilde{T}\|$ is the error in estimating the third order moment, $\epsilon_{\textnormal{pairs}}$ is the error in estimating the second order moments and $\sigma_{\min}$ is the $k^{\tha}$ singular value of the factor matrices.   While the $\epsilon_{\textnormal{pairs}}$ can be obtained by matrix Bernstein's bounds as before (e.g. see~\cite{AnandkumarHsuKakade:COLT12}), we have an improved bound for $\epsilon_{\textnormal{triples}}$ from Theorem~\ref{lem:TenConcentMixtureModel}, compared to previous results. Note that the first term corresponding to $\epsilon_{\textnormal{triples}}$  is the dominant one and we improve its scaling.
\end{remark}

\begin{remark}[RIP property] \label{remark:RIP}
Given $n$ samples for the model $\mathcal{S}$ proposed in Section~\ref{sec:multiview}, define noise matrix
\begin{equation} \label{eqn:noise matrix}
E_A : = [\veps^1_A,\veps^2_A,\dotsc,\veps^n_A] \in \Rbb^{d \times n},
\end{equation}
where $\veps^i_A \in \R^d$ is the $i$-th sample of noise vector $\veps_A$. $E_B$ and $E_C$ are similarly defined. These matrices need to satisfy the RIP property as follows which is adapted from \citet{candes2006near}.

\textit{\begin{enumerate}[label={(RIP)}]
\item \label{RIP} Matrix $E \in \Rbb^{d \times n}$ satisfies a weak RIP condition such that for any subset of $O\left(\frac{d}{\log^2 d}\right)$ number of columns, the spectral norm of $E$ restricted to those columns is bounded by $2$.
\end{enumerate}}

It is known that when $n = \poly(d)$, the above condition is satisfied with high probability for many random models such as when the entries are i.i.d. zero mean Gaussian or Bernoulli random variables.
\end{remark}

\paragraph{Proof ideas:}
The basic idea for proving the concentration result in Theorem~\ref{lem:TenConcentMixtureModel} is an $\veps$-net argument. We construct an $\veps$-net and then show that with high probability the norm of error tensor is bounded for every vector in the $\veps$-net.

In some cases even a usual $\eps$-net of size $e^{O(d)}$ is good enough. But, in many other cases the usual $\eps$-net construction does not provide a useful result since the failure probability is not small enough, and the union bound argument over all vectors in the $\veps$-net fails (or incurs additional polynomial factors in the sample complexity result). 
In particular, for a vector with high correlation with the data, we get a worse concentration bound. But, the key observation is that there can not be too many vectors that have high correlation with the data.
Therefore, for each fixed vector in the $\eps$-net, we partition the terms in the error into two sets; one set corresponds to the small terms (where the vector is not highly correlated with the data) and the other set corresponds to the large terms.
For the small terms, the usual $\eps$-net argument still works. For the large terms, we show that the number of such terms is limited. This is done either by RIP property of the noise matrices or by the bounded $2\to 3$ norm of factor matrices $A^\top$, $B^\top$ and $C^\top$. 
See the proofs of Claims~\ref{claim3}-\ref{claim1} for more details.
This partitioning argument is inspired by the entropy-concentration trade-off proposed in~\citep{RudelsonVershynin2009}; however, here we have a finer partitioning into several sets, while in \citep{RudelsonVershynin2009} the partitioning is done into only two sets.


\paragraph{Spherical Gaussian mixtures:}
Similar tensor concentration bound as above holds	 for the spherical Gaussian mixtures model with exploiting symmetrization trick as follows.
In the spherical Gaussian mixtures model, the modified higher order moment (tensor) in~\eqref{eqn:spherical Gaussian modified moment} is symmetric, and hence noise matrices $E_A$, $E_B$ and $E_C$ are all the same. This can cause a problem because some square terms in the error tensor are not zero mean and we need to show their concentration around the mean. The well-known {\em symmetrization technique} can be exploited here where we draw two independent set of samples, and show the difference between the two is with high probability small. This technique is widely applied to show concentration around the median, and in all our cases the median is very close to the mean.

\subsection{ICA and sparse ICA}

For the ICA model, we provide the tensor concentration result for the modified $4$th order observed moment (tensor) in~\eqref{eqn:ICA_modified moment} in both dense and sparse cases.

\begin{theorem}[Tensor concentration bound for ICA]
\label{lemma:ICA}
Consider $n$ samples $x^i = A h^i, i \in [n]$ from the ICA model with mixing matrix $A \in \R^{d \times k}$.
Suppose $\|A\| \le O(1+\sqrt{k/d})$ and the entries of $h \in \R^k$ are independent subgaussian variables with $\E[h_j^2]=1$ and constant nonzero  $4$th order cumulant.
For the  $4$th order cumulant $M_4$ in \eqref{eqn:ICA_modified moment} and its empirical estimate $\widehat{M}_4$, if $n\ge d$, we have with high probability
$$\|\widehat{M}_4 - M_4\| \le\tilde{O} \left( \frac{m^2}{n} + \sqrt{\frac{m^4}{d^3n}} \right),\quad m:=\max(d,k).$$
\end{theorem}

See the proof in Appendix~\ref{appendix:ICA concent bound}. We have an improved bound for the sparse ICA setting as follows.

\begin{theorem}[Tensor concentration bound for sparse overcomplete ICA]
\label{lemma:sparseICA}
In the ICA model $x=Ah$, suppose $h_j = s_jg_j$ where $s_j$'s are i.i.d. Bernoulli random variables with $\Pr [s_j=1] = s/k$, and $g_j$'s are independent $1$-subgaussian random variables. Consider $n$ independent samples $x^i = Ah^i, i \in [n],$ where each $h^i$ is distributed as $h$. Suppose $A$ satisfies~\ref{RIP} property (see Remark~\ref{remark:RIP} for details on RIP condition).
For  the $4$th order cumulant $M_4$ in \eqref{eqn:ICA_modified moment} and its empirical estimate $\widehat{M}_4$, if $n,k\ge d$, we have with high probability
$$\|\widehat{M}_4 - M_4\| \le \tilde{O} \left( \frac{s^2}{n} + \sqrt{\frac{s^4}{d^3n}} \right).$$
\end{theorem}

See the proof in Appendix~\ref{appendix:sparse ICA concent bound}.

{\em Dependence on $k$:}
It may seem counter-intuitive that the bound in Theorem~\ref{lemma:sparseICA} does not depend on $k$. The dependency on $k$ is actually in the expectation where the expected tensor $\E \bigl[ x^{\otimes 4} \bigr]$ in $M_4$ is close to $\frac{s}{k}\sum_{j \in [k]} a_j^{\otimes 4}$. We typically require the deviation to be less than the expected value.

\paragraph{Proof ideas:}
The proof ideas are similar to the multiview mixtures model where we provide $\veps$-net arguments and partition the terms to small and large ones.
In addition, for the ICA model, we exploit the subgaussian property of $h_j$'s to provide concentration bound for the summation of subgaussian random variables raised to the $4$th power (see Claim~\ref{claim:gaussian4th}). This implies the concentration bound for the $4$th order term $\E[x^{\otimes 4}]$ in $M_4$ (see Claim~\ref{claim:ICA4th}). For the $2$nd order term $T$ in $M_4$, the bound is argued using Matrix Bernstein's inequality (see Claim~\ref{claim:ICA2nd}).
For the sparse ICA model, the RIP property of $A$ is exploited to bound the size of intersection between the support of (partitioned) vectors in the $\veps$-net and the support of sparse vectors $h^i$ (see Claim~\ref{claim:sparseICA4th}). 



\section{Learning Algorithm}

In this section, we first introduce the tensor decomposition algorithm. Then, we provide some basic definitions and assumptions incorporated throughout the learning results. We conclude the section stating the organization of learning guarantees which are proposed in subsequent sections.

\subsection{Tensor decomposition algorithm} \label{sec:algorithm}
We exploit the tensor decomposition algorithm in~\citep{AltTensorDecomp2014} to learn the parameters of the latent variable models. This is given in Algorithm~\ref{algo:Power method form}.
The main step in~\eqref{eqn:asymmetric power update} basically performs alternating {\em asymmetric power updates}\,\footnote{This is exactly the generalization of asymmetric matrix power update to $3$rd order tensors.} on the different tensor modes.
Notice that the updates alternate among different modes of the tensor which can be viewed as a rank-$1$ form of the standard alternating least squares (ALS) method.
For vectors $v,w \in \R^d$, recall the definition of multilinear form $T(I,v,w) \in \R^d$ in \eqref{eqn:rank-1 update} where $T(I,v,w)$ is a multilinear combination of the tensor mode-$1$ fibers.

Intuition about the performance of tensor power update under non-orthogonal components is provided in~\citep{AltTensorDecomp2014}, which is reminded here.
For a rank-$k$ tensor $T$ as in~\eqref{eqn:tensordecomp}, suppose we start at the correct vectors $\ha=a_j$ and $\hb=b_j$,  for some $j \in [k]$. Then the numerator of tensor power update in~\eqref{eqn:asymmetric power update} is expanded as
\begin{equation}\label{eqn:intuition}
T \left( \ha, \hb, I \right)
= T \left( a_j, b_j, I \right)
= w_j c_j + \sum_{i \neq j} w_i \langle a_j,a_i \rangle \langle b_j,b_i \rangle c_i.
\end{equation}
We observe that under orthogonal components the second term is zero, and thus the true vectors $a_j,b_j$ and $c_j$ are stationary points for the power update procedure. However under incoherent (soft-orthogonal) components, the stationary points of the power update procedure are approximate estimates of the true components with small error.

The purpose of clustering step is to identify which initializations are successful in recovering the true components under unsupervised setting.
For more detailed discussion on the algorithm, see \citet{AltTensorDecomp2014}.

Notice that in this paper, the input tensor $T$ is the higher order moment of the LVMs described in Section ~\ref{sec:LVMs}. More details are stated in the learning results provided in next sections.

\begin{algorithm}[t]
\caption{Tensor decomposition via alternating power updates \citep{AltTensorDecomp2014}}
\label{algo:Power method form}
\begin{algorithmic}
\REQUIRE Tensor $T \in \Rbb^{d \times d \times d}$, number of initializations $L$, number of iterations $N$.
\FOR{$\tau=1$ \TO $L$}
\STATE \textbf{Initialize} unit vectors $\ha_\tau^{(0)} \in \Rbb^d$, $\hb_\tau^{(0)} \in \Rbb^d$, and $\hc_\tau^{(0)} \in \Rbb^d$ as
\bi[itemsep=-1mm]
\vspace{-2mm}
\item Semi-supervised setting: label information is exploited. See equation~\eqref{eqn:semi-supervised init}.
\item Unsupervised setting: SVD-based technique in Procedure~\ref{algo:SVD init} when $k \leq \beta d$ (for arbitrary constant $\beta$).
\ei
\FOR{$t=0$ \TO $N-1$}
\STATE Asymmetric power updates (see \eqref{eqn:rank-1 update} for the definition of the multilinear form):
\begin{align} \label{eqn:asymmetric power update}
\ha_\tau^{(t+1)} = \frac{T \left( I, \hb_\tau^{(t)}, \hc_\tau^{(t)} \right)}{\left\| T \left( I, \hb_\tau^{(t)}, \hc_\tau^{(t)} \right) \right\|}, \quad
\hb_\tau^{(t+1)} = \frac{T \left( \ha_\tau^{(t)}, I, \hc_\tau^{(t)} \right)}{\left\| T \left( \ha_\tau^{(t)}, I, \hc_\tau^{(t)} \right) \right\|}, \quad
\hc_\tau^{(t+1)} = \frac{T \left( \ha_\tau^{(t)}, \hb_\tau^{(t)},I \right)}{\left\| T \left( \ha_\tau^{(t)}, \hb_\tau^{(t)},I \right) \right\|}.
\end{align}
\ENDFOR
\STATE weight estimation:
\begin{align} \label{eqn:weight update}
\hw_\tau = T \left( \ha_\tau^{(N)}, \hb_\tau^{(N)}, \hc_\tau^{(N)} \right).
\end{align}
\ENDFOR
\STATE Cluster set $\left\{ \left( \hw_\tau,\ha_\tau^{(N)},\hb_\tau^{(N)},\hc_\tau^{(N)} \right), \tau \in [L] \right\}$ into $k$ clusters as in Procedure~\ref{alg:cluster}.
\RETURN the center member of these $k$ clusters as estimates $(\hw_j,\ha_j,\hb_j,\hc_j), j \in [k]$.
\end{algorithmic}
\end{algorithm}

\floatname{algorithm}{Procedure}
\begin{algorithm}[t]
\caption{Clustering process \citep{AltTensorDecomp2014}}
\label{alg:cluster}
\begin{algorithmic}
\REQUIRE Tensor $T \in \Rbb^{d \times d \times d}$, set of $4$-tuples 
$\left\{(\hw_\tau, \ha_\tau,\hb_\tau, \hc_\tau),\tau\in [L]\right\}$, parameter $\epsilon$.
\FOR{$i = 1$ \TO $k$}
\STATE Among the remaining 4-tuples, choose $\ha,\hb,\hc$ which correspond to the largest $|T(\ha,\hb,\hc)|$.
\STATE Do $N$ more iterations of alternating updates in \eqref{eqn:asymmetric power update} starting from $\ha,\hb,\hc$.
\STATE Let the output of iterations denoted by $(\ha,\hb,\hc)$ be the center of cluster $i$.
\STATE Remove all the tuples with $\max\{|\langle \ha_\tau,\ha\rangle|,|\langle \hb_\tau,\hb\rangle|,|\langle \hc_\tau,\hc\rangle|\} > \epsilon/2$.
\ENDFOR
\RETURN the $k$ cluster centers.
\end{algorithmic}
\end{algorithm}

\begin{algorithm}[t]
\caption{SVD-based initialization when $k = O(d)$ \citep{AltTensorDecomp2014}}
\label{algo:SVD init}
\begin{algorithmic}
\REQUIRE Tensor $T \in \Rbb^{d \times d \times d}$.
\STATE Draw a random standard Gaussian vector $\theta \sim \mathcal{N}(0,I_d).$
\STATE Compute $u_1$ and $v_1$ as the top left and right singular vectors of  $T(I,I,\theta) \in \R^{d \times d}$.
\STATE $\ha^{(0)} \leftarrow u_1$, $\hb^{(0)} \leftarrow v_1$.
\STATE Initialize $\hc^{(0)}$ by update formula in \eqref{eqn:asymmetric power update}.
\RETURN $\bigl( \ha^{(0)}, \hb^{(0)}, \hc^{(0)} \bigr)$.
\end{algorithmic}
\end{algorithm}

\paragraph{Efficient implementation given samples:} In Algorithm~\ref{algo:Power method form},  a given tensor $T$ is input, and we then perform the updates. However, in many settings (especially machine learning applications), the tensor is not available before hand, and needs to be computed from samples. Computing and storing the tensor can be enormously expensive for high-dimensional problems. Here, we provide a simple observation on how we can manipulate the samples directly to carry out the update procedure in Algorithm~\ref{algo:Power method form} as {\em multi-linear} operations, leading to efficient computational complexity.

Consider the mutiview mixtures model desribed in Section~\ref{sec:multiview} where the goal is to decompose the empirical moment tensor $\hat{T}$ of the form\begin{align} \label{eqn:empirical tensor}
\hT := \frac{1}{n} \sum_{l \in [n]} x_1^{(l)} \otimes x_2^{(l)} \otimes x_3^{(l)},
\end{align}
where $x_r^{(l)}$ is the $l^{\tha}$ sample from view $r \in [3]$.
Applying the power update \eqref{eqn:asymmetric power update} in Algorithm \ref{algo:Power method form} to $\hat{T}$, we have
\begin{equation}\label{eqn:multilinear}
\tl{c} 
:= \hT (\ha,\hb,I)
= \frac{1}{n} X_3 \bigl( X_1^\top \ha * X_2^\top \hb \bigr),
\end{equation}
where $*$ corresponds to the {\em Hadamard} product. Here, $X_r := \bigl[ x_r^{(1)} \ x_r^{(2)} \ \dotsb \ x_r^{(n)} \bigr] \in \Rbb^{d \times n}$. Thus, the update can be computed efficiently using simple matrix and vector operations.
It is easy to see that the above update in \eqref{eqn:multilinear} is easily parallelizable, and especially, the different initializations can be parallelized, making the algorithm scalable for large problems.

\subsection*{Basic definitions and assumptions}
The error bounds in the subsequent results are provided in terms of distance between the estimated and the true vectors. 
\begin{definition}
For any two vectors $u, v \in \R^d$, the {\em distance} between them is defined as
\begin{align} \label{eqn:dist function definition}
\dist(u,v) := \sup_{z \perp u} \frac{\langle z,v \rangle}{\| z \| \cdot \| v \|}
= \sup_{z \perp v} \frac{\langle z,u \rangle}{\| z \| \cdot \| u \|}.
\end{align}
\end{definition}

Note that distance function $\dist(u,v)$ is invariant w.r.t. norm of input vectors $u$ and $v$. Distance also provides an upper bound on the error between unit vectors $u$ and $v$ as (see Lemma A.1 of \citet{AgarwalEtal:SparseCoding2013})
\begin{align*}
\min_{z \in \{-1,1\}} \|zu-v \| \leq \sqrt{2} \dist(u,v).
\end{align*}
Incorporating distance notion resolves the sign ambiguity issue in recovering the components: note that a third order tensor is unchanged if the sign along one of the modes is fixed and the signs along the other two modes are flipped.

Here, we review some of the assumptions and settings assumed throughout the learning results provided in next sections. Consider tensor decomposition form in \eqref{eqn:tensordecomp}.
Let $A:=[a_1 \ a_2 \ \dotsb \ a_k] \in \Rbb^{d \times k}$ denote the {\em factor matrix}. Similar factor matrices are defined as $B$ and $C$ in the asymmetric cases, e.g., multiview linear mixtures model.
For simplicity and without loss of generality, we assume that the columns of factor matrices have unit $\ell_2$ norm, since we can always rescale them, and adjust the weights appropriately.
Also, for simplicity we assume $a_i,b_i,c_i \in \R^d, i \in [k],$ are uniformly i.i.d. drawn from the unit $d$-dimensional sphere $\Sc^{d-1}$ (see Remark~\ref{remark:generecity assump} for more details).

In this paper, we focus on learning in the challenging overcomplete regime where the number of components/mixtures is larger than observed dimension. Precisely, we assume $ k \geq \Omega(d)$.
Note that the results can be easily adapted to the highly undercomplete regime when $k \leq o(d)$.

\subsection*{Learning results organization}

In Section~\ref{sec:LVMs}, we described how learning different latent variable models can be formulated as a tensor decomposition problem by performing appropriate modifications on the observed moments. For those LVMs, the tensor concentration bounds are provided in Section~\ref{sec:Tensor Concent.}. We then proposed the tensor decomposition algorithm in Section~\ref{sec:algorithm} which is robust to noise.
Employing all these techniques and results, we finally provide learning results for different latent variable models including multiview linear mixtures, ICA and sparse ICA in the subsequent sections.
We consider two settings, \viz  semi-supervised setting, where a small amount of label information is available, and unsupervised setting where such information is not available.
In the former setting, we can handle overcomplete mixtures with number of components $k =o(d^{p/2})$, where $d$ is the observed dimension and $p$ is the order of observed moment. In the latter case, our analysis only works when $k \le \beta d$ for any constant $\beta$. See the following two sections for learning guarantees.



\section{Learning Multiview Linear Mixtures Model} \label{sec:multiview learning}

In this section, we provide the semi-supervised and unsupervised learning results for the multiview linear mixtures model described in Section~\ref{sec:multiview}.

\subsection{Semi-supervised learning}

In the semi-supervised setting, label information is exploited to build good initialization vectors for tensor decomposition Algorithm~\ref{algo:Power method form} as follows. For the multiview linear mixtures model in Figure~\ref{fig:Multiview}, let
$$x^{(l)}_{1,j}, x^{(l)}_{2,j}, x^{(l)}_{3,j} \in \Rbb^{d}, \quad j \in [k], l \in [m_j],$$
denote $m = \sum_{j \in [k]} m_j$ samples of vectors corresponding to different labels, where the samples with subscript $j$ have label $j$. Then, for any $j \in [k]$, we have the empirical estimate of mixture components as
\begin{align} \label{eqn:semi-supervised init}
\ha_j := \frac{1}{m_j} \sum_{l \in [m_j]} x^{(l)}_{1,j}, \quad
\hb_j := \frac{1}{m_j} \sum_{l \in [m_j]} x^{(l)}_{2,j}, \quad
\hc_j := \frac{1}{m_j} \sum_{l \in [m_j]} x^{(l)}_{3,j}.
\end{align}

Given $n$ unlabeled samples, let
\begin{equation} \label{eqn:target error mixture}
\epsilon_R :=
\left\{\begin{array}{ll}
\tl{O} \left( k \sqrt{d}/\sqrt{n} \right) + \tl{O} \left( \sqrt{k}/d \right), & \zeta^2 = \Theta(1), \\
\tl{O} \left( \sqrt{k/n} \right) + \tl{O} \left( \sqrt{k}/d \right), & \zeta^2 = \Theta \left( \frac{1}{d} \right),
\end{array}\right.
\end{equation}
denote the recovery error.
We first provide the settings of Algorithm~\ref{algo:Power method form} which include input tensor $T$, number of iterations $N$ and the initialization setting. 

\paragraph{Settings of Algorithm~\ref{algo:Power method form} in Theorem~\ref{thm:semi-supervised learning}:}
\bi[itemsep=-1mm]
\item Given $n$ unlabeled samples $x^{(i)}_{1}, x^{(i)}_{2}, x^{(i)}_{3} \in \Rbb^{d}, i \in [n]$, consider the empirical estimate of $3$rd order moment in \eqref{eqn:tensordecompMixture} as the input to Algorithm \ref{algo:Power method form}.
\item Number of iterations: $N = \Theta \left( \log \left( 1/\epsilon_R \right) \right)$.
\item Initialization: Exploit the empirical estimates in \eqref{eqn:semi-supervised init} as initialization vectors. 
\ei

\paragraph{Conditions for Theorem~\ref{thm:semi-supervised learning}:}
\bi[itemsep=-1mm]
\item Rank condition: $\Omega(d) \leq k \leq o(d^{3/2})$. 
\item The columns of factor matrices are uniformly i.i.d. drawn from unit $d$-dimensional sphere  $\Sc^{d-1}$ (see Remark~\ref{remark:generecity assump} for more discussion).
\item Suppose the distribution of observed variables given hidden state is sub-Gaussian, and the number of labeled samples with label $j$, denoted by  $m_j$, satisfies\,\footnote{In model $\Sc$, the columns of factor matrices are unit vectors, and therefore, the most reasonable regime of error is when the expected norm of error vector is constant, i.e., $\Ebb \left[ \|\zeta \sqrt{d} \veps\|^2 \right] = \zeta^2 d \leq O(1)$. But, note that the label complexity holds even if $\zeta^2 d \geq \omega(1)$.}
\begin{align} \label{eqn:samplesemisup}
m_j  \geq \tl{\Omega} \left( \zeta^2 d \right), \ j \in [k].
\end{align}
\item Given $n$ unlabeled samples, noise matrices $E_A$, $E_B$ and $E_C$ satisfy the RIP condition in \ref{RIP} which is satisfied with high probability for many random models (see Remark~\ref{remark:RIP} for details on RIP condition). The number of samples $n$ satisfies
\begin{align} \label{eqn:sample complexity MixtureModels}
n \geq
\left\{\begin{array}{ll}
\tl{\Omega} \left( k^2 d \right), & \zeta^2 = \Theta(1), \\
\tl{\Omega} \left( k \right), & \zeta^2 = \Theta \left( \frac{1}{d} \right),
\end{array}\right.
\end{align}
where $\zeta^2$ is the variance of each entry of observation vectors.
\ei

\begin{theorem}[Semi-supervised learning of multiview linear mixtures model] \label{thm:semi-supervised learning}
Assume the conditions and settings mentioned above hold.
Then, Algorithm \ref{algo:Power method form} outputs $\ha_j, j \in [k]$ as the estimates of columns of true factor matrix $A$ satisfying w.h.p.
\begin{align*}
\dist \left( \hat{a}_j, a_j \right) \leq \epsilon_R, \quad j \in [k],
\end{align*}
where $\dist(\cdot,\cdot)$ function and $\epsilon_R$ are defined in \eqref{eqn:dist function definition} and \eqref{eqn:target error mixture}, respectively.
Similar error bounds hold for other factor matrices $B$ and $C$.
In addition, the weight estimates $\hw_j, j \in [k]$ satisfy w.h.p.
\begin{align*}
\left| \hw_j - w_j \right| \leq  \epsilon_R/k, \quad j \in [k].
\end{align*}
\end{theorem}

See Appendix~\ref{appendix:proofs} for the proof.

{\em Approximation error in recovery:}
The recovery error $\epsilon_R$ involves two terms. One arises due to empirical estimation of $3$rd order moment (given by $\tl{O} \bigl( k \sqrt{d}/\sqrt{n} \bigr)$ or $\tl{O} \bigl( \sqrt{k/n} \bigr)$) and is inevitable. The other term is due to non-orthogonality of columns of factor matrices (given by $\tl{O} \bigl( \sqrt{k}/d \bigr)$) which is an approximation error in recovery of the tensor components. Note that the latter goes to zero for large enough $d$ since we have $k \leq o(d^{3/2})$.



\begin{remark}[Minimax sample complexity]
Note that the number of labeled samples required is much smaller than the number of unlabeled samples, i.e., $\sum_{j\in [k]} m_j \ll n$. 
Thus, we provide efficient learning guarantees for overcomplete multiview Gaussian mixtures in the semi-supervised setting under a small number of labeled samples.
Furthermore, in the low noise regime $\zeta^2 = \Theta \left( \frac{1}{d} \right)$, the sample complexity bounds for unlabeled samples is $\tl{\Omega}(k)$, which is the {\em minimax} bound  up to $\polylog$ factors.
\end{remark}

\begin{remark}[Different noise regime]
For brevity, both semi-supervised and unsupervised learning results for multiview linear mixtures model in this section are provided in low noise $\zeta^2 = \Theta (1/d)$ and high noise $\zeta^2 = \Theta(1)$ regimes. But, notice that the result for general regime of noise (all different magnitudes of $\zeta$) can be provided according to the general tensor concentration bound proposed in Theorem~\ref{lem:TenConcentMixtureModel}.
\end{remark}

\begin{remark}[Random assumption on factor matrices] \label{remark:generecity assump}
In the above learning result, we assume that the mixture components are uniformly i.i.d. drawn from unit $d$-dimensional sphere $\Sc^{d-1}$. This is a reasonable assumption for continuous models including the multiview linear mixtures model described here. But, it is not appropriate for discrete models where the non-negativity assumptions on the entries of factor matrices are required.
Moreover, the random assumption is provided for simplicity, while the original conditions for the guarantees of Algorithm~\ref{algo:Power method form} are deterministic; see~\citet{AltTensorDecomp2014}. They also show that random matrices satisfy these deterministic assumptions with high probability.
\end{remark}

\begin{remark}[Bounded $2 \to 3$ norm assumption]
Notice that the bounded $2 \to 3$ norm assumption in tensor concentration bound in Theorem~\ref{lem:TenConcentMixtureModel} is a weaker condition than assuming incoherence property for learning result in Theorem~\ref{thm:semi-supervised learning} which is needed for the algorithm guarantees. Furthermore, it is discussed in~\citet{AltTensorDecomp2014} that under the assumptions $k \leq o(d^{3/2})$ and uniform draws of columns of $A$, $B$ and $C$ from unit sphere, the bound on $2 \to 3$ norm is satisfied.
\end{remark}

\begin{remark}[Spherical Gaussian mixtures] \label{remark:spherical Gaussian}
Similar learning results as in Theorem~\ref{thm:semi-supervised learning} hold for the spherical Gaussian mixtures. It is discussed in Section~\ref{sec:SphericalGaussMix} how learning this model can be reduced to the tensor decomposition problem. Here, the $3$rd order empirical (modified) moment $\widehat{M}_3$ in \eqref{eqn:spherical Gaussian_empirical modified moment} is considered as the input of Algorithm \ref{algo:Power method form} with symmetric updates. Thus, we show minimax unlabeled sample complexity for semi-supervised learning of overcomplete spherical Gaussian mixtures.
\end{remark}

\subsection{Unsupervised learning}
In the unsupervised setting, there is no label information available to build the initialization vectors. Here, the initialization is performed by doing rank-$1$ SVD on random slices of the moment tensor proposed in Procedure~\ref{algo:SVD init}.
The conditions and settings for unsupervised learning are stated as follows where comparing to the semi-supervised learning, the initialization setting, rank and sample complexity conditions are changed.

\paragraph{Settings of Algorithm~\ref{algo:Power method form} in Theorem~\ref{thm:unsupervised learning}:}
\bi[itemsep=-1mm]
\item Given $n$ unlabeled samples $x^{(i)}_{1}, x^{(i)}_{2}, x^{(i)}_{3} \in \Rbb^{d}, i \in [n]$, consider the empirical estimate of $3$rd order moment in \eqref{eqn:tensordecompMixture} as the input to Algorithm \ref{algo:Power method form}.
\item Number of iterations: $N = \Theta \left( \log \left( 1 / \epsilon_R \right) \right)$.
\item The initialization in each run of Algorithm~\ref{algo:Power method form} is performed by SVD-based technique proposed in Procedure~\ref{algo:SVD init}, with the number of initializations as
$$L  \geq k^{\Omega \left( k^2/d^2 \right)}.$$
\ei

\paragraph{Conditions for Theorem~\ref{thm:unsupervised learning}:}
\bi[itemsep=-1mm]
\item Rank condition: $k = \Theta(d)$.
\item The columns of factor matrices are uniformly i.i.d. drawn from unit $d$-dimensional sphere  $\Sc^{d-1}$.
\item The number of samples $n$ satisfies
\begin{align*}
n \geq
\left\{\begin{array}{ll}
\tl{\Omega} \left( k^4 \right), & \zeta^2 = \Theta(1), \\
\tl{\Omega} \left( k^2 \right), & \zeta^2 = \Theta \left( \frac{1}{d} \right).
\end{array}\right.
\end{align*}
\ei


\begin{theorem}[Unsupervised learning of multiview linear mixtures model] \label{thm:unsupervised learning}
Assume the conditions and settings mentioned above hold.
Then, Algorithm \ref{algo:Power method form} outputs $\ha_j, j \in [k]$ as the estimates of columns of true factor matrix $A$ (up to permutation) satisfying w.h.p.
\begin{align*}
\dist \left( \hat{a}_j, a_j \right) \leq \epsilon_R, \quad j \in [k],
\end{align*}
where $\dist(\cdot,\cdot)$ function and $\epsilon_R$ are defined in \eqref{eqn:dist function definition} and \eqref{eqn:target error mixture}, respectively.
Similar error bounds hold for other factor matrices $B$ and $C$.
In addition, the weight estimates $\hw_j, j \in [k]$ satisfy w.h.p.
\begin{align*}
\left| \hw_j - w_j \right| \leq  \epsilon_R/k, \quad j \in [k].
\end{align*}
\end{theorem}

See Appendix~\ref{appendix:proofs} for the proof.

\begin{remark}[Comparison with ``whitening + moment-based" techniques in the undercomplete setting when $k \approx d$]
Here, we discuss how our approach makes a huge improvement on sample complexity for learning multiview linear mixtures model and spherical Gaussian mixtures with the additional {\em incoherence} property we assume.

{\em Multiview linear mixtures model:}  We compare with the previous result by~\citet{SongEtal:NonparametricTensorDecomp}, which employs whitening procedure followed by tensor power updates in the undercomplete setting. When $k \approx d$, the sample complexity in~\citep{SongEtal:NonparametricTensorDecomp} is scaled as $n \geq \tilde{\Omega}(k^{6.5})$.  In comparison, the sample complexity for our method scales as $\tl{\Omega}(k^2)$, which is far better. This is especially relevant in the high dimensional regime, where $k$ and $d$ are large, and our analysis shows lower sample complexity under incoherent factors.

{\em Spherical Gaussian mixtures:} As mentioned in Remark~\ref{remark:spherical Gaussian}, the above unsupervised learning result can be also adapted for learning mixture of spherical Gaussians. An algorithm for learning mixture of spherical Gaussians in the undercomplete setting is also provided in~\citep{SphericalGaussian2012}, which is a moment-based technique combined with a whitening step. When $k=d$, the sample complexity in~\citep{SphericalGaussian2012} scales as $n \geq \tl{\Omega}(k^3)$. But, our tight tensor concentration analysis leads to the better sample complexity of $n \geq \tl{\Omega}(k^2)$. Note that this comparison is in the low noise regime $\zeta^2 = \Theta \bigl( \frac{1}{d} \bigr)$.
\end{remark}


\section{Learning Independent Component Analysis (ICA)}
In this section, we propose the semi-supervised and unsupervised learning results for the ICA model described in Section~\ref{sec:ICA}. Unlike multi-view models, the standard ICA model does not have noise. This is because in ICA every sample is a mixture of many components (compared to multi-view models), and therefore, the noise is already ``built in'' the model.


\subsection{Semi-supervised learning}
By semi-supervised setting in ICA, we mean some prior information is available which provides good initializations for the components.

Given $n$ samples of observations $x^i = A h^i, i \in [n]$, let
\begin{equation} \label{eqn:target error ICA}
\tl{\epsilon}_R :=
\tl{O} \left( k^2 / \min \left\{ n,\sqrt{d^3 n} \right\} \right) + \tl{O} \left( \frac{\sqrt{k}}{d^{3/2}} \right)
\end{equation}
denote the recovery error.

\paragraph{Settings of Algorithm~\ref{algo:Power method form} in Theorem~\ref{thm:semi-supervised learning ICA}:}
\bi[itemsep=-1mm]
\item Given $n$ samples $x^i=Ah^i,  i \in [n]$, consider the empirical estimate of $4$th order (modified) moment $M_4$ (see \eqref{eqn:ICA_modified moment}) as the input to Algorithm \ref{algo:Power method form} with symmetric $4$th order updates. See~\citet{AltTensorDecomp2014} for higher order extension of the algorithm.
\item Number of iterations: $N = \Theta \left( \log \left( 1/ \tl{\epsilon}_R \right) \right)$.
\item Initialization: it is assumed that for any $j \in [k],$ an approximation of $a_j$ denoted by $\ha^{(0)}_j$ is given satisfying
$$\min_{z \in \{-1,1\}} \| z\ha^{(0)}_j - a_j \| \leq \frac{w_{\max}}{w_{\min}}.$$
Note that the initialization up to sign recovery is only required.
\ei

\paragraph{Conditions for Theorem~\ref{thm:semi-supervised learning ICA}:}
\bi[itemsep=-1mm]
\item Rank condition: $\Omega(d) \leq k \leq o(d^2)$.
\item The columns of factor matrices are uniformly i.i.d. drawn from unit $d$-dimensional sphere  $\Sc^{d-1}$.
\item The entries of $h$ are independent subgaussian variables with $\E[h_j^2]=1$ and constant nonzero  $4$th order cumulant.
\item The number of samples $n$ satisfies
$$
n \geq
\left\{\begin{array}{ll}
\tl{\Omega}(k^2), & k \leq O(d^{1.5})/\polylog(d), \\
\tl{\Omega} \left( k^4/d^3 \right), & \textnormal{o.w.}
\end{array}\right.
$$
\ei

\begin{theorem}[Semi-supervised learning of ICA] \label{thm:semi-supervised learning ICA}
Assume the conditions and settings mentioned above hold.
Then, Algorithm \ref{algo:Power method form} outputs $\ha_j, j \in [k]$ as the estimates of columns of true mixing matrix $A$ satisfying w.h.p.
\begin{align*}
\dist \left( \hat{a}_j, a_j \right) \leq \tl{\epsilon}_R, \quad j \in [k],
\end{align*}
where $\dist(\cdot,\cdot)$ function and $\tl{\epsilon}_R$ are defined in \eqref{eqn:dist function definition} and \eqref{eqn:target error ICA}, respectively.
In addition, the weight estimates $\hw_j, j \in [k]$ satisfy w.h.p.
\begin{align*}
\left| \hw_j - w_j \right| \leq  \tl{\epsilon}_R, \quad j \in [k].
\end{align*}
\end{theorem}

See Appendix~\ref{appendix:proofs} for the proof.

{\em Approximation error for recovery:}
Notice that the approximation error recovery for the ICA model is $\tl{O} \bigl(\sqrt{k}/d^{3/2} \bigr)$, while for the multiview mixtures model the approximation is $\tl{O} \bigl(\sqrt{k}/d \bigr)$. The difference is because of different tensor orders for the two models.

{\em Weight recovery comparison with multiview mixtures model:}
Comparing the weight recovery error for ICA model in Theorems~\ref{thm:semi-supervised learning ICA}~and~\ref{thm:unsupervised learning ICA} with the multiview linear mixtures model in Theorems~\ref{thm:semi-supervised learning}~and~\ref{thm:unsupervised learning}, we observe that the factor $1/k$ does not exist in the ICA model. This is because of different assumptions on the weights in the tensor form. For the multiview mixtures model, it is assumed $w_j = \Theta(1/k), j \in [k]$. But, in the ICA model, the weights are the 4th order cumulants $\kappa_j$ (see \eqref{eqn:ICA_tensor form}) which are assumed to be constant.

\begin{remark}[Efficient sample complexity]
We observe that for highly overcomplete regime $k = \Theta(d^2)/\polylog(d)$, the ICA model can be efficiently learned from fourth order moment with $n \geq \tl{\Omega}(k^{2.5})$ number of unlabeled samples. In the unsupervised setting, previous results require large polynomial sample complexity, e.g., \citet{fourierpca} need $\poly(k^4)$ number of unlabeled samples for learning ICA, where the poly factor is not explicitly characterized.
\end{remark}

\subsection{Unsupervised learning}
In the unsupervised setting, the initialization is performed by doing rank-$1$ SVD on random slices of the moment tensor proposed in Procedure~\ref{algo:SVD init}.
The conditions and settings for unsupervised learning are stated as follows where comparing to the semi-supervised learning, the initialization setting, rank and sample complexity conditions are changed.

\paragraph{Settings of Algorithm~\ref{algo:Power method form} in Theorem~\ref{thm:unsupervised learning ICA}:}
\bi[itemsep=-1mm]
\item Given $n$ samples $x^i=Ah^i,  i \in [n]$, consider the empirical estimate of $4$th order (modified) moment $M_4$ (see \eqref{eqn:ICA_modified moment}) as the input to Algorithm \ref{algo:Power method form} with symmetric $4$th order updates. See~\citet{AltTensorDecomp2014} for higher order extension of the algorithm.
\item Number of iterations: $N = \Theta \left( \log \left( 1 / \tl{\epsilon}_R \right) \right)$.
\item The initialization is performed by $4$-th order generalization\,\footnote{In the $4$th order case, the SVD is performed on $T(I,I,\theta,\theta) \in \R^{d \times d}$ for some random vector $\theta$.} of SVD-based technique in Procedure~\ref{algo:SVD init}, with the number of initializations as
$$L  \geq k^{\Omega ( k^2/d^2 )}.$$
\ei

\paragraph{Conditions for Theorem~\ref{thm:unsupervised learning ICA}:}
\bi[itemsep=-1mm]
\item Rank condition: $k = \Theta(d)$.
\item The columns of factor matrices are uniformly i.i.d. drawn from unit $d$-dimensional sphere  $\Sc^{d-1}$.
\item The entries of $h$ are independent subgaussian variables with $\E[h_j^2]=1$ and constant nonzero  $4$th order cumulant.
\item The number of samples $n$ satisfies
$$
n \geq \tl{\Omega} \left( k^3 \right).
$$
\ei

\begin{theorem}[Unsupervised learning of ICA] \label{thm:unsupervised learning ICA}
Assume the conditions and settings mentioned above hold.
Then, Algorithm \ref{algo:Power method form} outputs $\ha_j, j \in [k]$ as the estimates of columns of true mixing matrix $A$ (up to permutations) satisfying w.h.p.
\begin{align*}
\dist \left( \hat{a}_j, a_j \right) \leq \tl{\epsilon}_R, \quad j \in [k],
\end{align*}
where $\dist(\cdot,\cdot)$ function and $\tl{\epsilon}_R$ are defined in \eqref{eqn:dist function definition} and \eqref{eqn:target error ICA}, respectively.
In addition, the weight estimates $\hw_j, j \in [k]$ satisfy w.h.p.
\begin{align*}
\left| \hw_j - w_j \right| \leq  \tl{\epsilon}_R, \quad j \in [k].
\end{align*}
\end{theorem}

See Appendix~\ref{appendix:proofs} for the proof.

\subsection{Sparse ICA}

For the sparse ICA model introduced in Section~\ref{sec:ICA}, suppose the entries of $h$ are i.i.d. Bernoulli-subgaussian random entries, where the probability of each Bernoulli variable being $1$ is $s/k$, and therefore, $s$ is the expected number of nonzero entries in $h$.
More precisely, suppose $h_j = s_jg_j$ where $s_j$'s are i.i.d. Bernoulli random variables with $\Pr [s_j=1] = s/k$, and $g_j$'s are independent $1$-subgaussian random variables.

In the following theorem, we provide both semi-supervised and unsupervised learning of sparse ICA model. Note that the error recovery is changed as $$\tl{\epsilon}_R := \tl{O} \left( s k / \min \left\{ n,\sqrt{d^3 n} \right\} \right) + \tl{O} \left( \frac{\sqrt{k}}{d^{3/2}} \right).$$

\begin{theorem}[Semi-supervised and unsupervised learning of sparse ICA] \label{thm:learning sparse ICA}
Similar semi-supervised and unsupervised learning guarantees as in Theorems~\ref{thm:semi-supervised learning ICA}~and~\ref{thm:unsupervised learning ICA} hold for the sparse ICA model with the following sample complexity requirements.
For {\em semi-supervised} setting, we need
$$
n \geq
\left\{\begin{array}{ll}
\tl{\Omega}(sk), & sk \leq O(d^3)/\polylog(d), \\
\tl{\Omega} \left( s^2 k^2/d^3 \right), & \textnormal{o.w.},
\end{array}\right.
$$
and for {\em unsupervised} setting, we need
$$n \geq \tl{\Omega} \left( k^2 s \right).$$
In addition, here we assume that mixing matrix $A$ satisfies the RIP property in \ref{RIP} (see Remark~\ref{remark:RIP} for details on RIP condition).
\end{theorem}

See Appendix~\ref{appendix:proofs} for the proof.

\begin{remark}[Comparison with multiview mixtures and ICA]
In terms of sparsity of latent vector $h$, the sparse ICA spans between multiview Gaussian mixtures (where $h$ has one nonzero entry in vector representation), and ICA (where $h$ is fully dense). Comparing the guarantees, we also observe that the sample complexity results for sparse ICA bridges the range of models  between multiview mixtures model and ICA.
\end{remark}

{\em Comparison with previous approaches:}
The dictionary learning problem is also studied in~\citet{Arora2013,AgarwalEtal:SparseCoding2013,barak2014}. \citet{Arora2013,AgarwalEtal:SparseCoding2013} provide clustering based approaches for approximately learning incoherent dictionaries and then refining them through alternating minimization to obtain exact recovery of both the dictionary and the coefficients. They can handle sparsity level up to $O(\sqrt{d})$ (per sample) and the size of the dictionary $k$ can be arbitrary. \citet{barak2014} use the sum of squares framework and can handle the sparsity level up to (small enough) constant times $k$, but with the expense of computational complexity which scales as $k^{O(\log k)}$, and the size of the dictionary $k=O(d)$.
In addition, when the sparsity level is smaller as  $k^{1-\delta}$ for some $0<\delta<1$, their algorithm runs in polynomial time $k^{O(1/\delta)}$.
They can also go to higher level of overcompleteness with the expense of reducing sparsity level.
They do not need the assumptions that the dictionary is incoherent or that the coefficients are independent. They only have approximate recovery and note that exact recovery is impossible (from an identifiability standpoint) unless further assumptions are imposed. In contrast, we have a polynomial time method for incoherent dictionaries and independent coefficients which can handle arbitrary sparsity level, and provides approximate recovery. Moreover, we can handle larger dictionary sizes $k$ at the expense of more computation.

 Below, we show how we can extend our analysis to dependent sparsity setting, but with worse performance guarantees.

\subsubsection*{Extension to dependent sparsity}

In this section, we consider the noiseless sparse coding model $x = Ah$, but with no independence assumption on the   latent entries $h_i$'s. The analysis can be extended to noisy case.



We assume the following  moment conditions on $h$ in the dependent sparsity model. Note that these assumptions are comparable with the moment assumptions in~\citet{barak2014}.
\begin{align*}
\E \bigl[h_i^4 \bigr] = \E \bigl[h_i^2 \bigr] &= \beta s/k, \\
\E \bigl[h_i^2 h_j^2 \bigr] &\leq \tau, \quad i \neq j, \\
\E \bigl[h_i^3 h_j \bigr] & = 0, \quad i \neq j,
\end{align*}
with parameters $s$ and $\tau$, where $s$ is the expected number of nonzero entries in $h$, and $\beta$ is a universal constant. The first condition represents the normalization factor which depends on the sparsity level. The second condition limits the sparsity level and the amount of correlation between different entries of vector $h$. To provide more intuition about these parameters, assume that the entries of $h$ are distributed as Bernoulli-Gaussian random variables with each entry being nonzero with probability $s/k$. Then, we have $\tau = \rho p + (1-\rho)p^2$, where $\rho$ is the correlation coefficient between $h_i^2$ and $h_j^2$ for $i \neq j$.


\begin{theorem}[Noiseless sparse coding with dependent sparsity] \label{thm:sparse coding dependent}
Consider the described dictionary learning model $x=Ah$ where the moments of random vector $h$ satisfy the conditions stated before the theorem.
Let the noiseless $4$th order observed moment $\Ebb \bigl[ x^{\otimes 4} \bigr]$ be the input to Algorithm \ref{algo:Power method form} with symmetric 4th order updates.
Let the initialization in each run of Algorithm\ \ref{algo:Power method form} is performed by 4th order generalization of the SVD-based technique proposed in Procedure~\ref{algo:SVD init}. Let $\tl{\epsilon}_R := \tl{O} (\tau k/s) + \tl{O} \bigl( \sqrt{k}/d^{3/2} \bigr)$, and suppose
\begin{align*}
k = \Theta(d), \quad
N = \Theta \left( \log \left( 1/\tl{\epsilon}_R \right) \right),  \quad
L  \geq k^{\Omega \left( k^2/d^2 \right)}.
\end{align*}
In addition, assume that the columns of dictionary $A$ are uniformly i.i.d. drawn from unit $d$-dimensional sphere  $\Sc^{d-1}$.
If
\begin{align*}
\tau \leq \tl{O} \left( \frac{s/k}{d} \right),
\end{align*}
then whp
\begin{align*}
\dist \left( \hat{a}_j, a_j \right) \leq  \tl{\epsilon}_R, \quad j \in [k].
\end{align*}
\end{theorem}

See Appendix~\ref{appendix:proofs} for the proof.

Comparing with the dictionary learning result by~\citet{barak2014}, their algorithm is based on sum-of-squares techniques, and do not require any incoherence assumptions on the dictionary atoms. They can also handle higher levels of sparsity and correlation. On the other hand, they have a quasi-polynomial algorithm in the regime of high sparsity (small enough constant times $k$), while our algorithm is very simple and efficient.

The above analysis is in the noiseless regime, and the generalization to noisy case can be investigated as a future work which involves the sample complexity analysis in the dependent sparsity case. 



\section{Experiments}


In this Section, we run the algorithm for learning multiview Gaussian mixtures model. We consider model $\Sc$ described in Section~\ref{sec:multiview}. The mixture components are uniformly i.i.d. drawn from $d$-dimensional sphere $\Sc^{d-1}$. We assume low-noise regime such that $\zeta \sqrt{d} = 0.1$. In addition, let\,\footnote{In order to see the algorithm performance more easily, we generate $n$ samples such that each mixture component is exactly appeared in $\frac{n}{k}$ observations. Note that this is basically imposing equal number of different mixture components in the observations.} $w_j = \Pr[h=j] = \frac{1}{k}, j \in [k]$. We consider $d=100$ and $k= \{ 10, 20, 50, 100, 200, 500\}$. In order to see the effect of number of components $k$, we fix the number of samples $n=1000$.

Notice that the empirical tensor $\hT$ in~\eqref{eqn:empirical tensor} is not explicitly computed, and the tensor power updates in the algorithm are computed through the multilinear form stated in~\eqref{eqn:multilinear}. This leads to efficient computational complexity. See Section~\ref{sec:algorithm} for detailed discussion.

For each initialization $\tau \in [L]$, an alternative option of running the algorithm with a fixed number of iterations $N$ is to stop the iterations based on some stopping criteria. In this experiment, we stop the iterations when the improvement in subsequent steps is small as
\begin{align*}
\max \left( \left\| \ha_\tau^{(t)} - \ha_\tau^{(t-1)} \right\|^2, \left\| \hb_\tau^{(t)} - \hb_\tau^{(t-1)} \right\|^2, \left\| \hc_\tau^{(t)} - \hc_\tau^{(t-1)} \right\|^2  \right) \leq t_{\Stopping},
\end{align*}
where $t_{\Stopping}$ is the stopping threshold. According to the error bound provided in Theorem \ref{thm:semi-supervised learning}, we let
\begin{align} \label{eqn:stopping threshold}
t_{\Stopping} := t_1 (\log d)^2 \sqrt{\frac{k}{n}} + t_2 (\log d)^2 \frac{\sqrt{k}}{d},
\end{align}
for some constants $t_1, t_2>0$. Here, we set $t_1 = 1e-08$, and $t_2 = 1e-07$.

\begin{figure}[t]
\bc\bp
\psfrag{d eq 100, k eq 10}[l]{\scriptsize $d\!=\!100, k\!=\!10$}
\psfrag{d eq 100, k eq 20}[l]{\scriptsize $d\!=\!100, k\!=\!20$}
\psfrag{d eq 100, k eq 50}[l]{\scriptsize $d\!=\!100, k\!=\!50$}
\psfrag{d eq 100, k eq 100}[l]{\scriptsize $d\!=\!100, k=\!100$}
\psfrag{d eq 100, k eq 200}[l]{\scriptsize $d\!=\!100, k\!=\!200$}
\psfrag{d eq 100, k eq 500}[l]{\scriptsize $d\!=\!100, k\!=\!500$}
\psfrag{recovery rate of algorithm}[l]{\footnotesize recovery rate of algorithm}
\psfrag{number of initializations}[l]{\footnotesize number of initializations}
\psfrag{ratio of recovered columns}[l]{\footnotesize ratio of recovered components}
\includegraphics[width=3.5in]{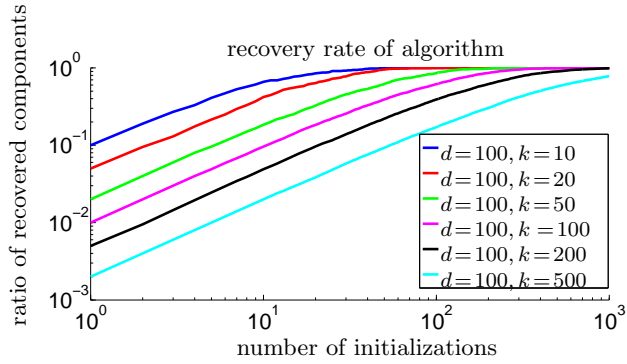}\ep\ec
\caption{\small Ratio of recovered components vs. the number of initializations. The figure is an average over 10 random runs.}
\label{fig:RecoveryRatio vs Init}
\end{figure}

A random initialization approach is used where $\ha^{(0)}$ and $\hb^{(0)}$ are uniformly i.i.d. drawn from sphere $\Sc^{d-1}$. Initialization vector $\hc^{(0)}$ is  generated through update formula in \eqref{eqn:asymmetric power update}.
Figure \ref{fig:RecoveryRatio vs Init} depicts the ratio of recovered components vs. the number of initializations. We observe that the algorithm is capable of recovering mixture components even in the overcomplete regime $k \geq d$.
As suggested in the experimental results of~\citet{AltTensorDecomp2014}, we also observe that random initialization works efficiently in the experiments, while the theoretical results for random initialization appear to be highly pessimistic. This suggests additional room for improving the theoretical guarantees under random initialization.

Table \ref{table:RecoveryRatio vs Init} provides the average square error of the estimates, the average weight error and the average number of iterations for different values of $k$.
The averages are over different initializations and random runs.
The square error is computed as
\begin{align*}
\frac{1}{3} \left[ \left\| a_j - \ha \right\|^2 + \left\| b_j - \hb \right\|^2 + \left\| c_j - \hc \right\|^2 \right],
\end{align*}
for the corresponding recovered column $j$. The weight error is computed as square relative error $|\hw - w_j |^2/w_j^2$. The number of iterations performed before stopping the algorithm is mentioned in the fourth column.
We observe that we can still get good error bounds even for overcomplete models with $d=100$ and $k=500$.

In the last two columns, the normalized values of errors are provided. The normalization is done by the number of mixtures $k$. Here, we observe that the normalized values (specially for the square error) are very close for different $k$. This complies with the theoretical error bound in~\eqref{eqn:target error  mixture} which claims that the square recovery error is bounded as $\tl{O}(k)$ when $d$ and $n$ are fixed as here.

\begin{table}[t]
\begin{center}
\caption{\small Results for learning a multi-view mixture model. $d=100$, $n=1000$, $\zeta\sqrt{d} = 0.1$.} \label{table:RecoveryRatio vs Init}
{\small \begin{tabular}{c||ccc|cc}
\hline
$k$ &
\begin{tabular}{c} avg. square \\ error \end{tabular} &
\begin{tabular}{c} avg. weight \\ error \end{tabular} &
\begin{tabular}{c} avg. \# of \\ iterations \end{tabular}  &
\begin{tabular}{c} avg. square \\ error $/k$ \end{tabular} &
\begin{tabular}{c} avg. weight \\ error $/k$ \end{tabular} \\
\hline \hline
10 & 1.24e-03 & 1.73e-05 & 9.81 & 1.24e-04 & 1.73e-06 \\
20 & 2.94e-03 & 5.28e-05 & 10.98 & 1.41e-04 & 2.64e-06 \\
50 & 7.21e-03 & 1.84e-04 & 12.74 & 1.44e-04 & 3.69e-06 \\
100 & 1.47e-02 & 5.36e-04 & 14.86 & 1.47e-04  & 5.36e-06 \\
200 & 3.03e-02 & 1.85e-03 & 18.34 & 1.51e-04 & 9.23e-06 \\
500 & 8.26e-02 & 1.23e-02 & 30.02 & 1.65e-04 & 2.45e-05 \\
\hline
\end{tabular}}
\end{center}
\end{table}

\subsubsection*{Acknowledgements}
A. Anandkumar is supported in part by Microsoft Faculty Fellowship, NSF Career award CCF-$1254106$, NSF Award CCF-$1219234$, and ARO YIP Award W$911$NF-$13$-$1$-$0084$. M. Janzamin is supported by  NSF Award CCF-1219234, ARO Award W911NF-12-1-0404  and ARO YIP Award W911NF-13-1-0084.

\renewcommand{\appendixpagename}{Appendix}

\appendixpage

\appendix

\section*{More Matrix  and Tensor Notations}

The outer product operator $\otimes$ defined earlier for vectors can be also generalized to higher order tensors. For instance, given matrices $A, B \in \R^{d \times d}$, the $4$th order tensor $T \in \R^{d \times d \times d \times d}$ is defined as
$$
T := A \otimes B \Leftrightarrow T_{i_1,i_2,i_3,i_4} = A_{i_1,i_2} B_{i_3,i_4}.
$$

For two matrices $A \in \Rbb^{d\times k}$ and $B\in \Rbb^{d\times k}$,  the {\em Hadamard} product is defined as the entry-wise multiplication of the matrices,
\[ A*B(i,j):= A(i,j) B(i,j), \quad i \in [d], j\in [k].\]

\section{Recap of Guarantees for Algorithm~\ref{algo:Power method form}} \label{appendix:guarantees}

In this section, we recap the local and global convergence guarantees of Algorithm~\ref{algo:Power method form} provided in \cite{AltTensorDecomp2014}. These results are required for proving unsupervised and semi-supervised learning results provided in this paper.

Let  $\psi := \|\Psi\|$ denote the spectral norm of error tensor $\Psi$, and
\begin{align} \label{eqn:target error}
\hat{\epsilon}_R := \frac{\psi}{w_{\min}} + \tl{O} \left( \gamma \frac{\sqrt{k}}{d} \right),
\end{align}
denote the recovery error where $\gamma := \frac{w_{\max}}{w_{\min}}$.

\subsection{Local convergence guarantee} \label{sec:local convergence}
The local convergence result is provided in the following theorem which bounds the estimation error after $N$ iterations of the Algorithm. Note that a good initialization is assumed in the local convergence guarantee and the behavior of asymmetric power update in the inner loop of Algorithm~\ref{algo:Power method form} is analyzed.

\paragraph{Settings of Algorithm~\ref{algo:Power method form} in Theorem~\ref{thm:local convergence}:}
\bi[itemsep=-1mm]
\item Number of iterations: $N = \Theta \left( \log \left( \frac{1}{\gamma \hat{\epsilon}_R} \right) \right)$, where $\gamma := \frac{w_{\max}}{w_{\min}}$.
\ei

\paragraph{Conditions for Theorem~\ref{thm:local convergence}:}
\bi[itemsep=-1mm]
\item Rank-$k$ true tensor with generic components: Let
\begin{align*}
T= \sum_{i\in [k]} \wstar_i \astar_i \otimes \bstar_i \otimes \cstar_i,\quad w_i>0, \astar_i, \bstar_i, \cstar_i \in \Sc^{d-1}, \forall\, i \in [k],
\end{align*}
where $a_i,b_i,c_i, i \in [k],$ are generated uniformly at random from the unit sphere $\Sc^{d-1}$.
\item Rank condition: $k = o \left( d^{1.5} \right).$
\item Perturbation tensor $\Psi$ satisfies the bound
$$\psi := \|\Psi\| \le \frac{w_{\min}}{6}.$$
\item Weight ratio: The maximum ratio of weights $\gamma := \frac{w_{\max}}{w_{\min}}$ satisfies the bound
\begin{align*}
\gamma = O \left( \min \left\{ \sqrt{d}, \frac{d^{1.5}}{k} \right\} \right).
\end{align*}
\item Initialization: The following initialization bound holds w.r.t. some $j \in [k]$ as
\begin{align} \label{eqn:good init}
\epsilon_0 := \max \left\{ \dist \left(\ha^{(0)}, a_j\right), \dist \left(\hb^{(0)}, b_j \right) \right\}
= O (1/\gamma),
\end{align}
where $\gamma := \frac{w_{\max}}{w_{\min}}$. In addition, given $\ha^{(0)}$ and $\hb^{(0)}$, suppose $\hc^{(0)}$ is also calculated by the update formula in \eqref{eqn:asymmetric power update}.
\ei

\begin{theorem}[Local convergence guarantee of Algorithm~\ref{algo:Power method form} \citep{AltTensorDecomp2014}] \label{thm:local convergence}
Consider $\hT = T + \Psi$ as the input to Algorithm \ref{algo:Power method form}, and assume the conditions and settings mentioned above hold.
Given initialization vectors $(\ha^{(0)}, \hb^{(0)}, \hc^{(0)})$, then the asymmetric power iterations (in the inner loop) of Algorithm \ref{algo:Power method form} satisfy the following bound w.h.p. after $N$ iterations as
\begin{align}\label{eqn:localconv}
\max \left\{ \dist \left( \ha^{(N)}, a_j \right), \dist \left( \hb^{(N)}, b_j \right), \dist \left( \hc^{(N)}, c_j \right) \right\} \leq O(\hat{\epsilon}_R),
\end{align}
where $\hat{\epsilon}_R$ is defined in \eqref{eqn:target error}.
Furthermore, the weight estimate $\hw = \hT \left( \ha^{(N)}, \hb^{(N)}, \hc^{(N)} \right)$ in \eqref{eqn:weight update} satisfies w.h.p.
\begin{align*}
\left| \hw - w_j \right| \leq O(w_{\min} \hat{\epsilon}_R). 
\end{align*}
\end{theorem}

Note that the recovery error $\hat{\epsilon}_R$ arises due to perturbation tensor $\Psi$ (given by $\frac{\psi}{w_{\min}} $) and non-orthogonality (given by $\tl{O} \left( \gamma \frac{\sqrt{k}}{d} \right)$). Thus, there is an approximation error in recovery of the tensor components. The above local convergence result can be also interpreted as an approximate local identifiability result for tensor decomposition under incoherent factors.

\subsection{Global convergence guarantee when $k=O(d)$} \label{sec:global convergence}

Theorem \ref{thm:local convergence} provides local convergence guarantee given good initialization vectors for different components. The global convergence guarantee is presented in the following theorem where  the SVD-based initialization method in Procedure~\ref{algo:SVD init} is exploited to provide good initialization vectors  when $k = O(d)$.

\paragraph{Settings of Algorithm~\ref{algo:Power method form} in Theorem~\ref{thm:global convergence}:}
\bi[itemsep=-1mm]
\item Number of iterations: $N = \Theta \left( \log \left( \frac{1}{\gamma \hat{\epsilon}_R} \right) \right)$, where $\gamma := \frac{w_{\max}}{w_{\min}}$.
\item The initialization in each run of Algorithm\ \ref{algo:Power method form} is performed by SVD-based technique proposed in Procedure~\ref{algo:SVD init}, with the number of initializations as
$$L  \geq k^{\Omega \left( \gamma^4 \left( k/d \right)^2 \right)}.$$
\ei

\paragraph{Conditions for Theorem~\ref{thm:global convergence}:}
\bi[itemsep=-1mm]
\item Rank-$k$ decomposition and perturbation conditions as\,\footnote{Note that the perturbation condition is stricter than the corresponding condition in the local convergence guarantee (Theorem~\ref{thm:local convergence}).}
\begin{align*}
T = \sum_{i\in[k]} w_i a_i\otimes b_i\otimes c_i, \quad \psi := \|\Psi\| \le \frac{w_{\min} \sqrt{\log k}}{\alpha_0 \sqrt{d}},
\end{align*}
where $a_i,b_i,c_i, i \in [k],$ are generated uniformly at random from the unit sphere $\Sc^{d-1}$, and $\alpha_0 >1$ is a constant.
\item Rank condition: $k = O(d).$
\ei

\begin{theorem}[Global convergence guarantee of Algorithm \ref{algo:Power method form} when $k=O(d)$, \citep{AltTensorDecomp2014}] \label{thm:global convergence}
Consider $\hT = T + \Psi$ as the input to Algorithm \ref{algo:Power method form}, and assume the conditions and settings mentioned above hold.
Then, for any $j \in [k]$, the output of Algorithm \ref{algo:Power method form} satisfies the following w.h.p.,
\begin{align*} 
\max \left\{ \dist \left( \ha_j, a_j \right), \dist \left( \hb_j, b_j \right), \dist \left( \hc_j, c_j \right) \right\}
& \leq O(\hat{\epsilon}_R), \\
\left| \hw_j - w_j \right| & \leq O(w_{\min} \hat{\epsilon}_R),
\end{align*}
where $\hat{\epsilon}_R$ is defined in \eqref{eqn:target error}.
\end{theorem}

The number of initialization trials $L$ is polynomial when $\gamma$ is a constant, and $k =O(d)$.

\section{Proof of Learning Theorems} \label{appendix:proofs}

The semi-supervised and unsupervised learning results for each latent variable model are proved by combining the corresponding tensor concentration bound proposed in Section~\ref{sec:Tensor Concent.} and the convergence guarantees of the tensor decomposition algorithm recapped in Appendix~\ref{appendix:guarantees}.

\bprfof{Theorem~\ref{thm:semi-supervised learning}}
The result is proved by applying the tensor concentration bound in Theorem~\ref{lem:TenConcentMixtureModel} to the local convergence result of Algorithm~\ref{algo:Power method form} recapped in Theorem~\ref{thm:local convergence}.
Note that in the high noise regime $\zeta^2 = \Theta(1)$, the term $\zeta^3 \sqrt{\frac{d}{n}}$ in Theorem~\ref{lem:TenConcentMixtureModel} is dominant, and in the low noise regime $\zeta^2 = \Theta \left( \frac{1}{d} \right)$, the term $\zeta \sqrt{w_{\max} \frac{d}{n}}$ in Theorem~\ref{lem:TenConcentMixtureModel} is dominant.

Note that the sub-Gaussian property of conditional observed distributions is used to provide the labeled sample complexity. Since the distribution of observed variables given hidden state is sub-Gaussian with covariance matrix $\zeta^2 I$ as in model $\Sc$ described in Section~\ref{sec:multiview}, we have the following
concentration bound where with probability at least $1-\delta$, the empirical estimate $\ha_j^{(0)}$ satisfies
$$
\left\| \ha_j^{(0)} - a_j \right\| \leq C_1 \sqrt{\frac{\zeta^2 d \log (1/\delta)}{m_j}}, \quad j \in [k],
$$
for some constant $C_1>0$.
\eprfof

\bprfof{Theorem~\ref{thm:unsupervised learning}}
The result is proved by applying the tensor concentration bound in Theorem~\ref{lem:TenConcentMixtureModel} to the global convergence result of Algorithm~\ref{algo:Power method form} recapped in Theorem~\ref{thm:global convergence}. The dominant error bounds in Theorem~\ref{lem:TenConcentMixtureModel} are the same as what stated in the proof of Theorem~\ref{thm:semi-supervised learning}.
\eprfof

\bprfof{Theorem~\ref{thm:semi-supervised learning ICA}}
The result is proved by applying the tensor concentration bound in Theorem~\ref{lemma:ICA} to the
local convergence result of Algorithm~\ref{algo:Power method form} in the $4$th order case. See~\citet{AltTensorDecomp2014} for the generalization of convergence result to higher order cases.
\eprfof

\bprfof{Theorem~\ref{thm:unsupervised learning ICA}}
The result is proved by applying the tensor concentration bound in Theorem~\ref{lemma:ICA} to the global convergence result of Algorithm~\ref{algo:Power method form} recapped in Theorem~\ref{thm:global convergence}.
Note that the SVD technique is applied to the $4$-th order case as described in the settings. Therefore, the requirement on noise in global convergence result is changed as
$\psi := \|\Psi\| \le \frac{w_{\min} \sqrt{\log k}}{\alpha_0^2 d}$.
\eprfof

\bprfof{Theorem~\ref{thm:learning sparse ICA}}
The learning results for the sparse ICA are proved similar to the ICA case, with the difference that the sparse ICA concentration bound in Theorem~\ref{lemma:sparseICA} is exploited here.
\eprfof

\bprfof{Theorem~\ref{thm:sparse coding dependent}}
Given linear model $x = Ah$, the $4$th order observed moment is expanded as
\begin{align} \label{eqn:Dictionary learning moment}
\Ebb \left[ x^{\otimes 4} \right] = \Ebb \left[ h^{\otimes 4} \right] \left( A^\top, A^\top, A^\top, A^\top \right),
\end{align}
where the multilinear notation defined in \eqref{eqn:multilinear form def} is exploited.

Expanding $\Ebb \left[ h^{\otimes 4} \right]$, and treating $\sum_{i \in [k]} \Ebb[h_i^4] \ e_i^{\otimes 4}$ as the main signal, the remaining term is
\begin{align*}
R := \sum_{i \neq j} \Ebb[h_i^2 h_j^2] \ e_i^{\otimes 2} \otimes e_j^{\otimes 2},
\end{align*}
where we also exploited the assumption that the expectation of terms involving odd powers of $h_i$ are zero.
Then, from \eqref{eqn:Dictionary learning moment}, the spectral norm of perturbation tensor is bounded as
\begin{align*}
\| \Psi \| := \left\| R \left( A^\top, A^\top, A^\top, A^\top \right) \right\|
= \left\| \sum_{i \neq j} \Ebb[h_i^2 h_j^2] \ a_i^{\otimes 2} \otimes a_j^{\otimes 2} \right\|
\leq \tau \|A\|^4,
\end{align*}
where we used $\Ebb[h_i^2 h_j^2] \leq \tau$ in the last inequality. Imposing condition $\| \Psi \| \leq \tl{O} \left( w_{\min}/ d \right)$, and then applying Theorem \ref{thm:global convergence}, the result is proved. Note that $w_{\min} := \min_{i \in [k]} \Ebb[h_i^4] = \beta s/k$.
\eprfof

\section{Proof of Tensor Concentration Bounds}
In this section, we provide the proof of tensor concentration bounds for different latent variable models including multiview linear mixtures model, ICA and sparse ICA.
In order to get polynomial sample complexity bounds for unlabeled samples in semi-supervised and unsupervised learning results, it is usually enough to treat the tensor as a vector/matrix and apply appropriate vector/matrix concentration bounds such as Bernstein bounds. However, these bounds can be significantly improved in many cases by considering the concentration property of the tensor spectral norm directly.

\subsection{Multiview linear mixtures model} \label{appendix:multiview concent bound}
In this section, we prove the tensor concentration result for the multiview linear mixtures model provided in Theorem~\ref{lem:TenConcentMixtureModel}.



\bprfof{Theorem~\ref{lem:TenConcentMixtureModel}}
Expanding the difference $\hat{T} - \tilde{T}$, we have
\begin{align}
\sublabon{equation}
\hat{T} - \tilde{T}
= & \ \frac{1}{n} \zeta^3 d^{1.5}  \sum_{i\in [n]} \veps^i_A \otimes \veps^i_B \otimes \veps^i_C \label{eqn:error3} \\
& + \frac{1}{n} \zeta^2  d \sum_{i\in [n]} \left( a_{h_i} \otimes \veps^i_B \otimes \veps^i_C + \veps^i_A \otimes b_{h_i} \otimes \veps^i_C + \veps^i_A \otimes \veps^i_B \otimes c_{h_i} \right) \label{eqn:error2} \\
& + \frac{1}{n} \zeta \sqrt{d} \sum_{i\in [n]} \left( a_{h_i} \otimes b_{h_i} \otimes \veps^i_C + a_{h_i} \otimes \veps^i_B \otimes c_{h_i} + \veps^i_A \otimes b_{h_i} \otimes c_{h_i} \right). \label{eqn:error1}
\end{align}
\sublaboff{equation}
There are three types of terms in the above difference which are bounded separately in Claims~\ref{claim3}-\ref{claim1} in Section~\ref{sec:claims}. Combining the results of claims, the theorem follows directly.

\eprfof

%


\subsubsection{Basic definitions and lemmata}

In the proof of the claims in Section~\ref{sec:claims}, we extensively apply two different types of partitioning as follows.

\begin{definition}[Small and large terms] \label{def:SmallLarge}
Consider matrices $E_A := [\veps^1_A,\veps^2_A,\dotsc,\veps^n_A] \in \R^{d\times n}$, and $E_B$ and $E_C$ which are similarly defined. For any set of vectors $u$, $v$ and $w$, the set of columns $[n]$ are partitioned into $2$ sets called sets of {\em small} and {\em large} terms according to the value of inner products $\inner{u,\veps^i_A}$, $\inner{v,\veps^i_B}$ and $\inner{w,\veps^i_C}$ as follows. The set of small values denoted by $L^c \subseteq [n]$ is defined as
\[ L^c := \left\{ i \in [n] :
|\inner{u,\veps^i_A}| < \frac{10\log d}{\sqrt{d}} \wedge
|\inner{v,\veps^i_B}| < \frac{10\log d}{\sqrt{d}} \wedge
|\inner{w,\veps^i_C}| < \frac{10\log d}{\sqrt{d}} \right\}, \]
and the rest of columns belong to the set of large values denoted by $L \subseteq [n]$.

Note that when necessary, the above partitioning is similarly applied to one or two matrices.
\end{definition}

\begin{lemma} \label{lem:NumLargeTerms}
Suppose matrix $E := [\veps^1,\veps^2,\dotsc,\veps^n] \in \R^{d\times n}$ satisfies the RIP property~\ref{RIP}. For a vector $u \in \R^d$, let set $L \subseteq [n]$ denote the set of columns of $E$ corresponding to large inner products $\inner{u,\veps^i}$ as defined in Definition~\ref{def:SmallLarge}, i.e.,
\[ L := \left\{ i \in [n] : |\inner{u,\veps^i}| \geq \frac{10\log d}{\sqrt{d}} \right\}. \]
Then, the size of set $L$ is bounded as
\begin{equation}  \label{eqn:NumLargeTerms}
|L| \leq \frac{d}{25\log^2 d}.
\end{equation}
\end{lemma}
\bprf
It can be shown by a contradiction argument assuming $|L| > \frac{d}{25\log^2 d}$. Consider submatrix $E[L]$ (matrix $E$ with columns restricted to set $L$). We have
\[ \|E\|^2 \geq \left\| E[L]^\top u \right\|^2 = \sum_{i \in L} \langle u, \veps^i \rangle^2 \ge |L| \frac{100\log^2 d}{d} > 4,\]
where the first inequality is from the definition of large terms for which $|\inner{u,\veps^i}| > 10\log d/\sqrt{d}$, and the second inequality is from contradiction assumption on $|L|$.
This contradicts with the RIP property that $\left\| E[L] \right\| \leq 2$, and therefore the bound in \eqref{eqn:NumLargeTerms} holds.
\eprf

The above partitioning into small and large sets is good when all we care about is the inner-products between a fixed vector and the noise vectors. However, when we are also interested in the inner-products between a fixed vector and columns of $A, B, C$, it is often not tight enough, and in order to get a tight bound, we propose the following finer partitioning.

\begin{definition}[Buckets and constrained vectors] \label{def:Buckets}
Consider matrix $C:=[c_1,c_2,\dotsc,c_k] \in \R^{d\times k}$, and let $t := \left\lceil \log_2 \sqrt{d} \right\rceil$. For any unit vector $w$, the set of columns $[k]$ are partitioned into $t+1$ buckets according to the value of inner products $\inner{c_j,w}$ as
\begin{align*}
K_0 &:= \left\{ j \in [k] : |\inner{c_j,w}| \leq \frac{1}{\sqrt{d}} \right\}, \\
K_l &:= \left\{ j \in [k] : |\inner{c_j,w}|\in \left( \frac{2^{l-1}}{\sqrt{d}}, \frac{2^l}{\sqrt{d}} \right] \right\}, \quad l \in [t].
\end{align*}

Furthermore, the constrained vector $z^l \in \R^k, l \in \{0,1,2,\dotsc,t\},$ corresponds to the inner products in bucket $l$ as
\[
z^l_j :=
\left\{\begin{array}{ll}
\inner{c_j,w}, & j \in K_l, \\
0, & j \notin K_l.
\end{array}\right.
\]
\end{definition}

One advantage of bucketing (which is not applicable to the small and large partitioning in the previous definition) is that buckets with large value has a smaller $\veps$-net. This exploits the additional property of matrices with bounded $2 \to 3$ norm.

\begin{lemma} \label{lem:buckets}
Consider matrix $C:=[c_1,c_2,\dotsc,c_k] \in \R^{d\times k}$ where the columns have unit norm, and $\|C^\top\|_{2 \to 3} = O(1)$. For a vector $w$ with unit norm, consider the buckets on columns of matrix $C$ defined in Definition~\ref{def:Buckets}. For constrained vector $z^l, l \in [t]$, let $p_l := 2^{l-1}$. Then, we have
\bi
\item $z^l$ has at most $O \left( \frac{d^{3/2}}{p_l^3} \right)$ nonzero entries.
\item There is an $\veps$-net of size
$\exp \left( O \left( \frac{d^{3/2}}{p_l^3} \left( \log k+\log \frac{1}{\veps} \right) \right) \right)$
for $z^l$.
\ei
\end{lemma}

\bprf
For the first part, we know the number of non-zero entries in $z^l$ is $|K_l|$. For any unit vector $w$, we have
\[
O(1) \ge \left\| C^\top w \right\|^3_3
\geq \sum_{j \in K_l} |\inner{w,c_j}|^3
\geq |K_l| \left( \frac{p_l}{\sqrt{d}} \right)^3,
\]
which implies the desired bound on $|K_l|$.


Let $q_l := O \bigl( \frac{d^{3/2}}{p_l^3} \bigr)$ be the maximum number of nonzero entries in $z^l$. 
First enumerate the support of $z^l$. There are ${k\choose q_l}$ possibilities for the location of $q_l$ nonzero entries in $z^l$ which is bounded as
$${k\choose q_l} \le \left( e \frac{k}{q_l} \right)^{q_l} \le e^{O \left( q_l \log k \right)}.$$
For a given support, take an $\veps$-net for all vectors in that support which has size
$$e^{O \left( q_l \log \left( \frac{1}{\veps} \right) \right)}.$$
The union of these $\veps$-nets is a valid $\veps$-net for $z^l$ of the desired size. This finishes the proof of second claim.

\eprf

A similar (but stronger) lemma can be proved for RIP matrices:

\begin{lemma}
\label{lem:bucketsRIP}
Consider matrix $E:=[\veps^1,\veps^2,\dotsc,\veps^n] \in \R^{d\times n}$ where the columns have unit norm, and it satisfies RIP property~\ref{RIP}. For a vector $w$ with unit norm, consider the buckets on columns of matrix $E$ defined in Definition~\ref{def:Buckets}. For constrained vector $z^l$, let $p_l := 2^{l-1}$. Then, for $l>4\log\log d$ we have
\bi
\item $z^l$ has at most $O \left( \frac{d}{p_l^2} \right)$ nonzero entries.
\item There is an $\veps$-net of size
$\exp \left( O \left( \frac{d}{p_l^2} \left( \log n+\log \frac{1}{\veps} \right) \right) \right)$
for $z^l$.
\ei
\end{lemma}

\bprf
The first claim follows from the same argument as in Lemma~\ref{lem:NumLargeTerms}. The $\veps$-net is constructed in the same way as in the previous lemma.
\eprf

\subsubsection{Proof of claims} \label{sec:claims}

In this section, we separately bound different error terms \eqref{eqn:error3}-\eqref{eqn:error1}.
Among all the terms, the terms like \eqref{eqn:error1} is most difficult to bound (intuitively because terms like $b_{h_i}$ are not ``as random'' as terms like $\veps_A^i$). In fact, the proof for the term \eqref{eqn:error1} can be adapted to bound all the other terms. Here for clarity we start from the simplest term \eqref{eqn:error3}, and point out new ideas in the proofs of \eqref{eqn:error2} and \eqref{eqn:error1}.

\begin{claim}[Bounding norm of \eqref{eqn:error3}] \label{claim3}
With high probability over $\veps_A^i,\veps_B^i,\veps_C^i$'s and $h_i$'s, we have
$$\left\| \frac{1}{n} \sum_{i=1}^n \veps^i_A\otimes \veps^i_B\otimes \veps^i_C \right\| \le \tilde{O} \left( \frac{1}{n} + \frac{1}{d\sqrt{n}} \right).$$
\end{claim}

\bprf
Let
$$T_1 := \frac{1}{n} \sum_{i=1}^n \veps^i_A\otimes \veps^i_B \otimes \veps^i_C.$$
Rewrite the tensor as
\begin{align} \label{eqn:T_1}
T_1 = \frac{1}{n} \sum_{i=1}^n \eta_i\veps^i_A\otimes \veps^i_B \otimes \veps^i_C,
\end{align}
where $\eta_i$'s are independent random $\pm 1$ variables with $\Pr[\eta_i = 1] = 1/2$. Clearly, $T_1$ has the same distribution as the original term, because of the symmetry in error vectors implying e.g. $\eta_i\veps^i_A \sim \veps^i_A$.
We first sample the vectors $\veps^i_A,\veps^i_B,\veps^i_C$, and therefore, the remaining random variables are just the $\eta_i$'s.

The goal is to bound norm of $T_1$ in \eqref{eqn:T_1} which is defined as
\begin{equation} \label{eqn:tensorconcen1}
\|T_1\| := \sup_{\|u\| = \|v\| = \|w\| = 1} | T_1(u,v,w) | =  \sup_{\|u\| = \|v\| = \|w\| = 1} \left| \frac{1}{n} \sum_{i=1}^n \eta_i \inner{u,\veps^i_A} \inner{v,\veps^i_B} \inner{w,\veps^i_C} \right|.
\end{equation}

In order to bound the above, we provide an $\eps$-net argument.
Construct an $\eps$-net for vectors $u$, $v$ and $w$ with $\eps = 1/n^2$.
By standard construction, size of the $\eps$-net is $e^{O(d\log n)}$.
First, for any fixed triple $(u,v,w)$, we bound $|T_1(u,v,w)|$ where $T_1(u,v,w)$ is a sum of independent variables. As introduced in Definition~\ref{def:SmallLarge}, we partition the sum into {\em large} and {\em small} terms as
\[ T_1(u,v,w) = \frac{1}{n} \sum_{i=1}^n \eta_i\inner{u,\veps^i_A} \inner{v, \veps^i_B} \inner{w, \veps^i_C} := S_L + S_{L^c},\]
where $S_{L^c}$ is the sum of {\em small} terms consisting of terms satisfying
$$\left\{
|\inner{u,\veps^i_A}| < \frac{10\log d}{\sqrt{d}} \wedge
|\inner{v,\veps^i_B}| < \frac{10\log d}{\sqrt{d}} \wedge
|\inner{w,\veps^i_C}| < \frac{10\log d}{\sqrt{d}} \right\},$$
and $S_L$ is the sum of  {\em large} terms including all the other terms.

{\em Bounding $|S_{L^c}|$}:
The sum $S_{L^c}$ is just a weighted sum of $\eta_i$'s, and the Bernstein's Inequality is exploited to bound it. Each term in the summation is bounded as
\begin{align*}
\left| \frac{1}{n} \inner{u,\veps^i_A} \inner{v,\veps^i_B} \inner{w,\veps^i_C} \right|
\leq O \left( \frac{\log^3 d}{n d^{3/2}} \right),
\end{align*}
where the bound on the small terms is exploited. The variance term is also bounded as
\begin{align*}
O \left( \frac{\log^6 d}{n d^3} \right).
\end{align*}
Applying Bernstein's inequality, with probability at least  $1 - e^{-Cd\log n}$ (where $C$ is a large enough constant), the sum of small terms $|S_{L^c}|$ is bounded by $\tilde{O} \left( \frac{1}{d\sqrt{n}} \right)$.

{\em Bounding $|S_L|$}:
From RIP property~\ref{RIP}, we know that noise matrices $E_A : = [\veps^1_A,\dotsc,\veps^n_A]$, $E_B : = [\veps^1_B,\dotsc,\veps^n_B]$ and $E_C:=[\veps^1_C,\dotsc,\veps^n_C]$ satisfy the weak RIP condition with high probability such that for any subset of $O \left( \frac{d}{\log^2 d} \right)$ number of columns, the spectral norm of matrices restricted to those columns is bounded by $2$.
Let $L$ denote the set of large terms in the proposed partitioning, and $E_A[L]$, $E_B[L]$ and $E_C[L]$ be the matrices $E_A$, $E_B$ and $E_C$ restricted to the columns indexed by $L$. Applying Lemma~\ref{lem:NumLargeTerms}, we have
\begin{equation*}
|L| \leq \frac{3d}{25\log^2 d}.
\end{equation*}
Note that an additional factor $3$ shows up here since the set of small terms is defined as the intersection of $3$ sets comparing to what proved in Lemma~\ref{lem:NumLargeTerms}.
Therefore, RIP property of $E_A$, $E_B$ and $E_C$ implies that $E_A[L]$, $E_B[L]$ and $E_C[L]$ have spectral norm bounded by $2$.
Now applying triangle inequality, we have
$$|S_L| \leq \frac{1}{n}\sum_{i\in L} |\inner{u,\veps^i_A}| \cdot |\inner{v,\veps^i_B}|\cdot |\inner{w,\veps^i_C}|\le \frac{1}{n}\sum_{i\in L} |\inner{u,\veps^i_A}| \cdot |\inner{v,\veps^i_B}| \le \frac{1}{n} \left\|E_A[L]^\top u\right\| \cdot \left\|E_B[L]^\top v \right\| \le \frac{4}{n},$$
where the second step uses the fact that $|\inner{w,\veps^i_C}| \le 1$, the third step exploits Cauchy-Schwartz inequality, and the last step uses bounds $\|E_A[L]\| \leq 2$ and $\|E_B[L]\| \leq 2$.
Notice the three matrices are already sampled before we do the $\veps$-net argument, and therefore, we do not need to do union bound over all $u,v,w$ for this event.

At this point, we have bounds on $|S_L|$ and $|S_{L^c}|$ for a fixed triple $(u,v,w)$ in the $\eps$-net. By applying union bound on all vectors in the $\eps$-net, the bound holds for every triple $(u,v,w)$ in the $\eps$-net. The argument for other $(u,v,w)$'s which are not in the $\eps$-net follows from their closest triples in the $\eps$-net.
\eprf

\begin{claim}[Bounding norm of \eqref{eqn:error2}] \label{claim2}
With high probability over $\veps_A^i,\veps_B^i$'s and $h_i$'s, we have
$$\left\| \frac{1}{n} \sum_{i=1}^n \veps^i_A\otimes \veps^i_B\otimes c_{h_i} \right\| \le \tilde{O} \left( \frac{1}{n} + \sqrt{\frac{w_{\max}}{n\sqrt{d}}} \right).$$
\end{claim}

\bprf
The proof is similar to the previous claim. Let
$$T_2 = \frac{1}{n} \sum_{i=1}^n \eta_i\veps^i_A\otimes \veps^i_B\otimes c_{h_i},$$
where $\eta_i$'s are independent random $\pm 1$ variables  with $\Pr[\eta_i = 1] = 1/2$.
Similar to the previous claim, we first sample the vectors $\veps^i_A,\veps^i_B$ and $h_i$'s, and therefore, the remaining random variables are just the $\eta_i$'s.
Assume the matrices $E_A$, $E_B$ satisfy the RIP property, and the number of times $h_i = j$ for $j\in [k]$ is bounded by $[nw_{\min}/2,2nw_{\max}]$.
All the events happen with high probability when $n \ge \tilde{\Omega}(1/w_{\min})$ and $n\le \poly(k)$.

The goal is to bound $\|T_2\|$. We construct an $\eps$-net for vectors $u$ and $v$ with $\eps = 1/n^2$. First, for any fixed pair $(u,v)$, we bound $\|T_2(u,v,I)\|$ where $T_2(u,v,I)$ is a sum of independent zero mean vectors. As introduced in Definition~\ref{def:SmallLarge}, consider partitioning on columns of $E_A$ and $E_B$ as
\[ T_2(u,v,I) = \frac{1}{n} \sum_{i=1}^n \eta_i \inner{u,\veps^i_A} \inner{v, \veps^i_B} c_{h_i} = S_L + S_{L^c},\]
where $S_{L^c}$ is the sum of {\em small} terms consisting of terms satisfying
$$\left\{ |\inner{u,\veps^i_A}| < \frac{10\log d}{\sqrt{d}} \ \wedge \ |\inner{v,\veps^i_B}| < \frac{10\log d}{\sqrt{d}} \right\},$$
and $S_L$ is the sum of {\em large} terms including all the other terms.

{\em Bounding $\|S_L\|$}:
This is bounded in a similar way to the argument for bounding $S_L$ in the previous claim.
From RIP property~\ref{RIP}, we know that noise matrices $E_A : = [\veps^1_A,\dotsc,\veps^n_A]$ and $E_B : = [\veps^1_B,\dotsc,\veps^n_B]$ satisfy the weak RIP condition with high probability.
Let $L$ be the set of large terms in the proposed partitioning, and $E_A[L]$, $E_B[L]$ be the matrices $E_A$, $E_B$ restricted to the columns indexed by $L$.
Applying Lemma~\ref{lem:NumLargeTerms}, we have
\begin{equation*}
|L| \leq \frac{2d}{25\log^2 d}.
\end{equation*}
Therefore, RIP property of $E_A$ and $E_B$ implies that $E_A[L]$ and $E_B[L]$ have spectral norm bounded by $2$.
Applying triangle inequality, we have
$$\|S_L\| \leq \frac{1}{n}\sum_{i\in L} |\inner{u,\veps^i_A}| \cdot |\inner{v,\veps^i_B}| \le \frac{1}{n} \left\|E_A[L]^\top u\right\| \cdot \left\|E_B[L]^\top v \right\| \le \frac{4}{n},$$
where Cauchy-Schwartz inequality is exploited in the second inequality, and the bounds $\|E_A[L]\| \leq 2$ and $\|E_B[L]\| \leq 2$ are used in the last inequality. Notice the two matrices are already sampled before we do the $\veps$-net argument, and therefore, we do not need to do union bound over all $u,v$ for this event.

{\em Bounding $\|S_{L^c}\|$}:
Similar to how we bounded $|S_{L^c}|$ in the previous claim by applying Bernstein's inequality, it is tempting to apply vector Bernstein's inequality here.
However, vector Bernstein's inequality does not utilize the fact that the matrix $C^\top$ has small $2\to 3$ norm, and results in a suboptimal bound. Here, we try to exploit this additional property to to get a better bound.

Let $L^c$ denote the set of small terms in the proposed partitioning on columns of $E_A$ and $E_B$. Then, we have
\begin{align*}
\inner{S_{L^c}, w} = \frac{1}{n} \sum_{i \in L^c} \eta_i \inner{u,\veps^i_A} \inner{v, \veps^i_B} \inner{w,c_{h_i}}.
\end{align*}
Now, we try to bound the above inner product $\inner{S_{L^c}, w}$ by considering an $\veps$-net on $w$ as well (Note that the $\veps$-nets on $u$ and $v$ are already considered). To do that we partition the inner products $\inner{c_j,w}$ into $t+1$ buckets ($t:=\lceil \log_2 \sqrt{d}\rceil$) as defined in Definition~\ref{def:Buckets} where
\begin{align*}
K_0 &:= \left\{ j \in [k] : |\inner{c_j,w}| \leq \frac{1}{\sqrt{d}} \right\}, \\
K_l &:= \left\{ j \in [k] : |\inner{c_j,w}|\in \left( \frac{2^{l-1}}{\sqrt{d}}, \frac{2^l}{\sqrt{d}} \right] \right\}, \quad l \in [t].
\end{align*}
Let $Q_l$ denote the sum of all terms that fall into bucket $K_l$ as
\begin{align} \label{eqn:Q_l}
Q_l := \frac{1}{n}  \sum_{i \in L^c, h_i \in K_l} \eta_i \inner{u,\veps^i_A}\inner{v,\veps^i_B}\inner{w,c_{h_i}}.
\end{align}
Note that by construction of buckets, we have $$\inner{S_{L^c},w} = \sum_{l=0}^t Q_l.$$
There are only $O(\log d)$ terms in this summation, and therefore, it suffices to show each term $Q_l$ is small.

For $Q_0$, it is a weighted sum of $\eta_i$'s with weights bounded by $\tilde{O}(1/d^{3/2})$, so the situation is exactly the same as Claim~\ref{claim3}.

For $Q_l, l \in [t]$, the argument is as follows. Let $p_l := 2^{l-1}$. Applying Lemma~\ref{lem:buckets}, we have
\[ |K_l| \leq O \left( \frac{d^{3/2}}{p_l^3} \right).\]
As stated in the beginning of proof, each hidden state $h_i \in [k]$ appears in at most $O(2n w_{\max})$ samples w.h.p. Hence, the total number of terms in the summation form \eqref{eqn:Q_l} for $Q_l$ is w.h.p. bounded as
\[ \left| \{ i \in [n]: h_i \in K_l \} \right| \leq O\left( n w_{\max} \frac{d^{3/2}}{p_l^3} \right). \]
Now the sum $Q_l$ in \eqref{eqn:Q_l} is a weighted sum of $\eta_i$'s and the Bernstein's inequality is exploited to bound it. Each term in the summation is bounded as
$$\tilde{O} \left( \frac{p_l}{n d^{3/2}} \right),$$
where the bound on the small terms and the bound on terms in bucket $K_l$ are exploited. The variance term is also bounded as
$$O \left(\frac{w_{\max}}{np_l d^{3/2}} \right).$$
Applying Bernstein's inequality, with probability at least $1 - e^{-Cd\log n}$ for large enough constant $C$, we have (notice below that $p_l\le O(\sqrt{d})$)
$$Q_l \leq \tilde{O} \left( \frac{p_l}{\sqrt{d}n} + \sqrt{\frac{w_{\max}}{np_l \sqrt{d}}} \right)
\le \tilde{O} \left( \frac{1}{n} + \sqrt{\frac{w_{\max}}{n\sqrt{d}}} \right).$$

At this point, we have bounds on $\|S_L\|$ and $\|S_{L^c}\|$ for a fixed pair of vectors $(u,v)$ in the $\eps$-net. By applying union bound on all vectors in the $\eps$-net, the bound holds for every pair $(u,v)$ in the $\eps$-net. The argument for other $(u,v)$'s which are not in the $\eps$-net follows from their closest pairs in the $\eps$-net.
\eprf

Now we are ready to bound the last term \eqref{eqn:error1}.

\begin{claim}[Bounding norm of \eqref{eqn:error1}] \label{claim1}
With high probability over $\veps_A^i$'s and $h_i$'s,  we have
\[ \left\| \frac{1}{n} \sum_{i=1}^n \veps^i_A\otimes b_{h_i}\otimes c_{h_i} \right\| \le
\tilde{O} \left(\frac{1}{n} + \sqrt{\frac{w_{\max}}{n}}\right).\]
\end{claim}

\bprf
Again, rewrite the tensor as
\begin{align} \label{eqn:T_3}
T_3 = \frac{1}{n} \sum_{i=1}^n \eta_i \veps^i_A\otimes b_{h_i}\otimes c_{h_i},
\end{align}
where $\eta_i$'s are independent random $\pm 1$ variables with $\Pr[\eta_i = 1] = 1/2$.
First sample $\veps^i_A$ and $h_i$'s, and therefore, the remaining random variables are just the $\eta_i$'s. In addition, assume $E_A := [\veps^1_A, \veps^2_A, \dotsc, \veps^n_A]$ satisfies the RIP property~\ref{RIP} and each $h_i \in [k]$ appears between $nw_{\min}/2$ and $2nw_{\max}$ times where both events happen with high probability.

The goal is to bound norm of $T_3$ in \eqref{eqn:T_3} which is defined as
\begin{equation} \label{eqn:tensorconcen3}
\|T_3\| := \sup_{\|u\| = \|v\| = \|w\| = 1} | T_3(u,v,w) | =  \sup_{\|u\| = \|v\| = \|w\| = 1} \left| \frac{1}{n} \sum_{i=1}^n \inner{u,\veps^i_A} \inner{v,b_{h_i}} \inner{w,c_{h_i}} \right|.
\end{equation}
In order to bound the above, we provide an $\eps$-net argument similar to what we did for bounding $S_{L^c}$ in the previous claim with the difference that here we apply bucketing to all three matrices $E_A$, $B$ and $C$.  First, for any fixed triple $(u,v,w)$, we partition the inner products in \eqref{eqn:tensorconcen3} into buckets as defined in Definition~\ref{def:Buckets}. Let $K^a_l$,  $K^b_l$ and $K^c_l$ denote the bucketing of matrices $E_A$, $B$ and $C$, respectively.

In addition, we merge the buckets $K^a_0$, $K^a_1,\dotsc,K^a_{4\log \log d}$ into $K^a_0$. This means $K^a_0$ now contains all $i$'s with inner product
$$|\inner{\veps^i_A,u}| \le \frac{16\log d}{\sqrt{d}},$$ and $K^a_l$'s for $1 \le l \le 4\log\log d$ are empty.
Let
$$ J_{l_1,l_2,l_3} := \left\{ i \in [n] : i \in K^a_{l_1} \wedge h_i \in K^b_{l_2} \wedge h_i \in K^c_{l_3} \right\},$$
and $Q_{l_1,l_2,l_3}$ be the sum of terms in summation \eqref{eqn:tensorconcen3} on this set, i.e.,
\begin{align} \label{eqn:Q_i3}
Q_{l_1,l_2,l_3} := \frac{1}{n} \sum_{i \in J_{l_1,l_2,l_3}} \inner{u,\veps^i_A} \inner{v,b_{h_i}} \inner{w,c_{h_i}}.
\end{align}
Note that by construction of buckets, the summation in \eqref{eqn:tensorconcen3} is expanded as
$$\frac{1}{n} \sum_{i=1}^n \inner{u,\veps^i_A} \inner{v,b_{h_i}} \inner{w,c_{h_i}} = \sum_{l_1,l_2,l_3=0}^t Q_{l_1,l_2,l_3}.$$
There are only $O(t^3) = O(\log^3 d)$ terms in this summation, and therefore, it suffices to show each term $Q_{l_1,l_2,l_3}$ is small.

For $Q_{0,0,0}$, it is a weighted sum of $\eta_i$'s with weights bounded by $\tilde{O}(1/d^{3/2})$, and therefore, it follows from the same arguments as Claim~\ref{claim3}.

For $Q_{l_1,l_2,l_3}$ with $\max\{l_1,l_2,l_3\} > 0$, let $p_l := 2^{\max\{l_1,l_2,l_3\}-1}$. By Lemma~\ref{lem:buckets} and Lemma~\ref{lem:bucketsRIP}, the total number of terms in the summation form \eqref{eqn:Q_i3} for $Q_{l_1,l_2,l_3}$ is w.h.p. bounded as
\[ \left| J_{l_1,l_2,l_3} \right| \leq O\left( n w_{\max} \frac{d^{3/2}}{p_l^3} \right),\]
and there exists an $\veps$-net of size
$$\exp \left( O\left( \frac{d^{3/2}}{p_l^3} \log n \right) \right)$$
with $\veps < 1/n^2$.
For every $u,v,w$ in the $\veps$-net, this term $n\cdot Q_{l_1,l_2,l_3}$ is a weighted sum of $\eta_i$'s, and the Bernstein's inequality is exploited to bound it.
Each term in the summation is bounded as $\frac{8p_l^3}{d^{3/2}},$ where the bound on the terms in buckets are exploited. The variance term is also bounded as
$$O \left( n w_{\max} \frac{p_l^3}{d^{3/2}} \right).$$
Applying Bernstein's inequality, with probability at least $1 - \exp \left( -C \frac{d^{3/2}}{p_l^3}\log n \right)$ for large enough constant $C$, we have
$$nQ_{l_1,l_2,l_3} \leq \tilde{O} \left( 1 + \sqrt{nw_{\max}} \right).$$
Taking the union bound over all triples in $\veps$-net, this bound holds for all such triples. For $u,v,w$ which are not in the $\veps$-net, the bound follows from the closest point in the $\veps$-net.

\eprf

\subsection{ICA} \label{appendix:ICA concent bound}

In this section, we prove the tensor concentration result for the ICA model provided in Theorem~\ref{lemma:ICA}.

Recall the $4$th order modified moment tensor in equation~\eqref{eqn:ICA_modified moment} as
$$M_4 := \Ebb [x \otimes x \otimes x \otimes x] - T, $$
where $T \in \R^{d \times d \times d \times d}$ is the fourth order tensor with
$$ T_{i_1,i_2,i_3,i_4} := \Ebb[x_{i_1} x_{i_2}] \Ebb[x_{i_3} x_{i_4}] + \Ebb[x_{i_1} x_{i_3}] \Ebb[x_{i_2} x_{i_4}] + \Ebb[x_{i_1} x_{i_4}] \Ebb[x_{i_2} x_{i_3}], \quad i_1,i_2,i_3,i_4 \in [d]. $$
Let $\widehat{M}_4$ be the empirical estimate of $M_4$ given $n$ samples.

\bprfof{Theorem~\ref{lemma:ICA}}
Let $W := \frac{1}{n}\sum_{i=1}^n x^i (x^i)^\top$, and therefore, the empirical estimate of $T$ is given by
\begin{equation} \label{eqn:ICA_2nd order term empirical}
\widehat{T}_{i_1,i_2,i_3,i_4} = W_{i_1,i_2}W_{i_3,i_4}+W_{i_1,i_3}W_{i_2,i_4}+W_{i_1,i_4}W_{i_2,i_3}.
\end{equation}
Then, the empirical estimate of $M_4$ is given by $$\widehat{M}_4 = \frac{1}{n}\sum_{i=1}^n (x^i)^{\otimes 4} - \widehat{T}.$$ The proof directly follows from Claims~\ref{claim:ICA4th} and \ref{claim:ICA2nd}, which bound the perturbation of the two terms separately. Claim~\ref{claim:ICA4th} bounds the $4$th order term perturbation $\Ebb [x^{\otimes 4}] - \frac{1}{n}\sum_{i=1}^n (x^i)^{\otimes 4}$, and Claim~\ref{claim:ICA2nd} bounds the 2nd order term perturbation $T - \widehat{T}$.
\eprfof

\subsubsection{Proof of claims}
Before bounding the 4-th order term we first give the following claim which bounds a sum of subgaussian variables raised to the $4$-th power.

\begin{claim}
\label{claim:gaussian4th}
Suppose $h_i, i \in [n]$, are independent $q$-subgaussian random variables. Then, for any $d \ge 1$, with probability at least $1- e^{-\omega(d\log n)}$ we have
$$\left| \frac{1}{n} \sum_{i=1}^n \left( h_i^4-\E \left[ h_i^4 \right] \right) \right| \le \tilde{O} \left( \frac{q^4d^2}{n} +\sqrt{\frac{q^8d}{n}} \right).$$
\end{claim}

(Notice that here $d$ is intended to be the dimension in later applications. However, for this claim we can choose $d$ to be an arbitrary real number that is at least $1$.)

\bprf
We prove
\begin{equation} \label{eqn:MedianBound}
\Pr \left[  \frac{1}{n} \left| \sum_{i=1}^n h_i^4 - \mbox{med} \Bigl( \sum_{i=1}^n h_i^4 \Bigr) \right| \le \tilde{O}\left( \frac{q^4d^2}{n} + \sqrt{\frac{q^8d}{n}} \right) \right] \geq  1 - e^{-\omega(d\log n)},
\end{equation}
where $\mbox{med}(\cdot)$ is the median of the distribution. By doing simple integration (for $d$ from $1$ to $\infty$), this concentration bound implies 
$$\left|\E \left[ \sum_{i=1}^n h_i^4\right] - \mbox{med} \Bigl( \sum_{i=1}^n h_i^4 \Bigr) \right|\le \tilde{O} \left( \frac{q^4}{\sqrt{n}} \right).$$
Therefore, when $d \ge 1$ the difference between mean and median is negligible, and we get the desired bound in the claim.

In order to prove the deviation bound from the median in~\eqref{eqn:MedianBound}, we use the standard symmetrization argument: it is enough to take two independent sample sets $\{h_1, h_2,\dotsc,h_n\}$ and $\{\tl{h}_1, \tl{h}_2,\dotsc,\tl{h}_n\}$ with the same distribution, and bound $\bigl| \frac{1}{n} \sum_{i\in [n]} \bigl( h_i^4 - \tl{h}_i^4 \bigr) \bigr|$. In order to bound the sum, we rewrite it in the form
$$
Q = \frac{1}{n} \sum_{i\in [n]} \eta_i |h_i^4 - \tl{h}_i^4|,
$$
where $\eta_i$'s are independent random $\pm 1$ variables with $\Pr[\eta_i = 1] = 1/2$.

Now we partition the terms in the summation for $Q$ into multiple buckets according to the magnitude of $\bigl| h_i^4 - \tl{h}_i^4 \bigr|$. Let $t:=\lceil \log_2 d^2 + C' \log_2\log_2 n \rceil$ (where $C'$ is a large enough constant). Then the buckets are defined as
\begin{align*}
K_0 &:= \left\{ i \in [n] : |h_i^4 - \tl{h}_i^4| \leq q^4 \right\}, \\
K_l &:= \left\{ i \in [n] : |h_i^4 - \tl{h}_i^4|\in \left( 2^{l-1} q^4 , 2^l q^4  \right]  \right\}, \quad l \in [t], \\
K_{t+1} & := \left\{ i \in [n] : |h_i^4 - \tl{h}_i^4| > 2^t q^4  \right\}.
\end{align*}
Let $Q_l$ denote the sum of all terms that fall into bucket $K_l$ as
\begin{align} 
Q_l := \frac{1}{n}  \sum_{i\in [n], i \in K_l}  |h_i^4 - \tl{h}_i^4| \eta_i.
\end{align}
Note that by construction of buckets, the original summation $Q = \sum_{l=0}^{t+1} Q_l.$
There are only $O(\log d)$ terms in this summation, and therefore, it suffices to show each term $Q_l$ is small.

Note that since $h_i$'s and $\tl{h}_i$'s are $q$-subgaussian random variables, we have
\begin{align}
\Pr \left[|h_i^4 -\tl{h}_i^4| \geq \lambda q^4 \right] & \leq
\Pr \left[h_i^4 \geq \lambda q^4/2 \right] + \Pr \left[\tl{h}_i^4 \geq \lambda q^4/2 \right] \nn \\
& = 2\Pr \left[ |h_i| \geq (\lambda q^4/2)^{1/4} \right] \nn \\
& \leq 4 \exp \left( - \frac{\sqrt{\lambda}}{2 \sqrt{2}} \right), \label{eqn:SubGaussBound}
\end{align}
where the last inequality uses $q$-subgaussian property.

For $Q_l, 0\le l \le 2\log \log n$, we apply Bernstein's inequality directly. Each term in the summation for $Q_l$ is bounded as $\tilde{O}(q^4/n)$, and the variance term is also bounded as $\tilde{O}(q^8/n)$. By applying Bernstein's inequality, with probability at least $1 - e^{-\omega(d\log n)}$, we have
$$Q_l \leq \tl{O} \left( \frac{q^4d}{n} + \sqrt{\frac{q^8 d}{n}} \right), \quad 0\le l \le 2\log \log n.$$

For $Q_l, 2\log \log n < l \le t$, we first bound the number of terms in bucket $K_l$. From \eqref{eqn:SubGaussBound}, we have 
$$\Pr \left[ |K_l| \geq \tilde{\Omega} \left(d2^{-l/2}\right) \right] \leq e^{-\omega(d\log n)}.$$
Each term in the summation $Q_l$ is bounded by $2^l q^4/n$, and therefore, by applying triangle inequality we have with probability at least $1 - e^{-\omega(d\log n)}$
$$
Q_l \le \tilde{O} \left( d2^{-l/2} \right) \frac{2^l q^4}{n} \le \tilde{O} \left( \frac{q^4 d 2^{l/2}}{n} \right) \le \tilde{O} \left(\frac{q^4d^2}{n} \right), \quad 2\log \log n < l \le t.
$$
Here the last inequality uses the fact that $l\le t$, which implies $2^{l/2} = \tilde{O}(d)$.

For the last term $Q_{t+1}$, again from~\eqref{eqn:SubGaussBound}, we have with probability at least $1-e^{-\omega(d\log n)}$, there is only one term in the sum and that particular term is smaller than $\tilde{O}(q^4d^2/n)$.

Now by union bound, with probability at least $1-e^{-\omega(d\log n)}$ all the terms are bounded by $\tilde{O} \bigl(q^4d^2/n+\sqrt{q^8d/n}\bigr)$, which implies the summation $Q$ is also bounded by
$$\tilde{O} \left( \frac{q^4d^2}{n}+\sqrt{\frac{q^8d}{n}} \right).$$
\eprf

Now we are ready to bound the 4-th order term perturbation $\Ebb [x^{\otimes 4}] - \frac{1}{n}\sum_{i=1}^n (x^i)^{\otimes 4}$.

\begin{claim}
\label{claim:ICA4th}
Suppose $\|A\| \le O(\sqrt{k/d})$ and the entries of $h \in \R^k$ are independent subgaussian variables with $\E[h_j^2]=1$.
Given $n$ samples $x^i = A h^i, i \in [n]$, we have with high probability
$$\left\| \frac{1}{n}\sum_{i \in [n]} (x^i)^{\otimes 4} - \E[x^{\otimes 4}] \right\| \le \tilde{O} \left( \frac{k^2}{n} + \sqrt{\frac{k^4}{d^3n}} \right).$$
\end{claim}

\bprf
The desired spectral norm in the lemma is defined as
$$\sup_{\|u\|=1} \left| \frac{1}{n} \sum_{i \in [n]} \inner{u,x^i}^4 - \E[\inner{u,x}^4] \right|.$$
In order to bound it, we provide an $\eps$-net argument.
Construct an $\eps$-net for vectors $u$ in the unit ball $\Sc^{d-1}$ with $\eps = 1/n^2$. By standard construction, size of the $\eps$-net is $e^{O(d\log n)}$. For any fixed $u$ in the $\eps$-net, let $v := A^\top u$. Since $x^i = Ah^i, i\in [n],$ we have $\inner{u,x^i} = \inner{v,h^i}$. Therefore, for any fixed $u$ (and the corresponding $v$) in the $\eps$-net, we would like to bound
$$Q := \frac{1}{n} \sum_{i \in [n]} \left( \inner{v,h^i}^4 - \E[\inner{v,h^i}^4] \right).$$
Since $h^i$'s have independent subgaussian entries, we know that $\inner{v,h^i}$ is $\|v\|$-subgaussian. On the other hand, we have
$$\|v\| \leq \|A\|\|u\| = O(\sqrt{k/d}),$$ and therefore, $\inner{v,h^i}$ is a $O(\sqrt{k/d})$-subgaussian random variable. By Claim~\ref{claim:gaussian4th},  with probability at least $1 - e^{-Cd\log n}$ (for large enough constant $C$) we have
$$|Q| \le \tilde{O} \left( \frac{k^2}{n}+\sqrt{\frac{k^4}{d^3n}} \right).$$
By applying union bound on all vectors in the $\eps$-net, the bound holds for every vector $u$ in the $\eps$-net.
The argument for other $u$'s which are not in the $\eps$-net follows from their closest vectors in the $\eps$-net.

\eprf

The 2nd order term $T$ in~\eqref{eqn:ICA_2nd order term} is sum of three terms, each of which is an outer-product of two matrices. Hence, it is good enough to apply a matrix concentration for bounding this term.

\begin{claim}
\label{claim:ICA2nd}
Suppose $\|A\| \le O(\sqrt{k/d})$ and the entries of $h \in \R^k$ are independent subgaussian variables with $\E[h_j^2]=1$.
Given $n$ samples $x^i = Ah^i, i \in [n]$, for $T$ in~\eqref{eqn:ICA_2nd order term} and the empirical estimate $\widehat{T}$ in~\eqref{eqn:ICA_2nd order term empirical}, if $n\ge d$, we have with high probability
$$\|\widehat{T} - T\| \le\tilde{O} \left(\sqrt{\frac{k^4}{d^3n}} \right).$$
\end{claim}

\bprf
Recall $W := \frac{1}{n}\sum_{i=1}^n x^i (x^i)^\top$.
We prove the result for the first term
$$\widehat{T}_1[i_1,i_2,i_3,i_4] = W_{i_1,i_2}W_{i_3,i_4},$$
or equivalently $\widehat{T}_1 = W\otimes W$. The analysis for the other two terms follow similarly from symmetry.

Let $T_1 = \E[xx^\top]\otimes \E[xx^\top] = \E[W]\otimes \E[W]$. We have
$$\widehat{T}_1-T_1 = (W-\E[W])\otimes \E[W]+\E[W]\otimes (W-\E[W])+(W-\E[W])\otimes (W-\E[W]).$$
For any matrices $A$ and $B$, we have $\|A\otimes B\| \le \|A\|\|B\|$. Thus,
\begin{equation}
\|\widehat{T}_1-T_1\| \le 2\|W-\E[W]\| \cdot \|\E[W]\|+\|W-\E[W]\|^2.\label{eqn:outerproduct}
\end{equation}
We bound $\|W-\E[W]\|$ by Matrix Bernstein's inequality. For applying Matrix Bernstein's inequality, we need a bound on the norm of each term in the summation form of $W$, i.e., bound on $\|x^i (x^i)^\top\|$ which holds almost surely. Therefore, we apply the Bernstein's inequality on the bounded version of $W$ as
$$W' := \frac{1}{n}\sum_{i=1}^n x^i(x^i)^\top \mathbf{1}_{\|x^i\| \le O(\sqrt{k}\log n)},$$
where $\mathbf{1}_{\|x^i\| \le O(\sqrt{k}\log n)}$ is an indicator variable. 
Since $x = Ah$ and entries of $h$ are subgaussian, the indicator variables are 1 with probability $1-n^{-\log n}$. Therefore, $W$ and $W'$ are equal with high probability at it suffices to apply Matrix Bernstein's bound on $W'$.

For the summation $W'$, the norm of each term is bounded by $\tilde{O}(k/n)$, and for the variance term, we have
$$\E \left[ W'(W')^\top \right] = \frac{1}{n}\E \left[ \|x^i\|^2x^i(x^i)^\top\mathbf{1}_{\|x^i\| \le O(\sqrt{k}\log n)} \right] \preceq \frac{1}{n}\tilde{O}(k)\E \left[ x^i(x^i)^\top \right] = \frac{1}{n}\tilde{O}(k)AA^\top.$$
Since $\|A\| \le O(\sqrt{k/d})$, it is concluded that the variance is bounded by $\tilde{O}(k^2/dn)$. Therefore, Matrix Bernstein's inequality implies that with probability at least $1-d/n$,
$$\|W'-\E[W']\| \le \tilde{O} \left( \frac{k}{n} + \frac{k}{\sqrt{dn}} \right).$$
Since $W$ is equal to $W'$ with high probability and $\|\E[W]-\E[W']\|$ is negligible, we also have $\|W-\E[W]\| \le \tilde{O}(k/\sqrt{dn})$ (when $n\ge d$).

On the other hand, $\E[W] = AA^\top$, and therefore, $\|\E[W]\| \leq k/d$. From \eqref{eqn:outerproduct}, we have $$\|\widehat{T}_1-T_1\| \le\tilde{O} \left(\sqrt{\frac{k^4}{d^3n}} \right).$$
\eprf

\subsection{Sparse ICA} \label{appendix:sparse ICA concent bound}

In this section, we prove the tensor concentration result for the sparse ICA model provided in Theorem~\ref{lemma:sparseICA}. This is the sparse coding problem in the sparse ICA setting (where $h_i$'s are independent and sparse).
The proof can be generalized to the case when $h_i$'s are negatively correlated or more generally when concentration bounds hold for $h_i$'s.

The proof of Theorem~\ref{lemma:sparseICA} is similar to the proof of Theorem~\ref{lemma:ICA}, where the $4$th order term perturbation $\Ebb [x^{\otimes 4}] - \frac{1}{n}\sum_{i=1}^n (x^i)^{\otimes 4}$, and the 2nd order term perturbation $T - \widehat{T}$ are separately bounded in the following two claims.
First, we bound the perturbation of the $4$th order term in the following claim. Note that this is the sparse version of Claim~\ref{claim:ICA4th}.

\begin{claim} \label{claim:sparseICA4th}
Consider the sparse ICA model described in Theorem~\ref{lemma:sparseICA}.
Given $n$ independent samples $x^i = Ah^i, i \in [n],$ we have with high probability
$$\left\| \frac{1}{n}\sum_{i=1}^n (x^i)^{\otimes 4} - \E[x^{\otimes 4}] \right\| \le \tilde{O} \left( \frac{s^2}{n} + \sqrt{\frac{s^4}{d^3n}} \right).$$
\end{claim}

\bprf
The proof uses ideas from both Claims~\ref{claim2} and \ref{claim:gaussian4th} . Without loss of generality, we assume $s/k < 1/2$. Otherwise, $h_j$'s are $2$-subgaussian, and therefore the dense case argument in Claim~\ref{claim:ICA4th} implies the desired bound.

Let $\eta_i$'s be independent random $\pm 1$ variables with $\Pr[\eta_i = 1] = 1/2$.
We equivalently bound
$$ \left\|  \frac{1}{n}\sum_{i=1}^n \eta_i \left( (x^i)^{\otimes 4} - \E \left[(x^i)^{\otimes 4} \right] \right) \right\| :=
\sup_{\|u\|=1} \left| \frac{1}{n} \sum_{i \in [n]} \eta_i \left( \inner{u,x^i}^4 - \E[\inner{u,x^i}^4] \right) \right|.$$
In order to bound it, we provide an $\eps$-net argument.
Construct an $\eps$-net for vectors $u$ in the unit ball $\Sc^{d-1}$ with $\eps = 1/n^2$. By standard construction, size of the $\eps$-net is $e^{O(d\log n)}$. For any fixed $u$ in the $\eps$-net, let $v := A^\top u$. Since $x^i = Ah^i, i\in [n],$ we have $\inner{u,x^i} = \inner{v,h^i}$. Therefore, for any fixed $u$ (and the corresponding $v$) in the $\eps$-net, we would like to bound
\begin{equation} \label{eqn:SparseICA proof1}
\left| \frac{1}{n} \sum_{i \in [n]} \eta_i \left( \inner{v,h^i}^4 - \E[\inner{v,h^i}^4] \right) \right|.
\end{equation}
Now, we follow the ideas of Claim~\ref{claim:gaussian4th}, and apply the standard symmetrization trick:
it is enough to take two independent sample sets $\{h^1, h^2,\dotsc,h^n\}$ and $\{\tl{h}^1, \tl{h}^2,\dotsc,\tl{h}^n\}$ with the same distribution, and bound $\bigl| \frac{1}{n} \sum_{i\in [n]} \eta_i \bigl( \inner{v,h^i}^4 - \inner{v,\tl{h}^i}^4 \bigr) \bigr|$ instead of~\eqref{eqn:SparseICA proof1}. 
Note that the difference between mean and median here is negligible because our distributions have  first and second moments polynomial in parameters, and strong exponential concentration. Therefore, for any vector $u$ (and the corresponding $v$), we would like to bound the sum
$$
\frac{1}{n} \sum_{i\in [n]} \eta_i \left| \inner{v,h^i}^4-\inner{v,\tl{h}^i}^4 \right|.
$$
The techniques we use to prove bounds on sums of random variables $\sum_{i=1}^n \eta_i z_i$ (either Bernstein's inequality, or bounding the number of terms and then using triangle inequality) all works if we just know an {\em upper bound} of $z_i$. Therefore, we can equivalently bound
$$
Q = \frac{1}{n} \sum_{i\in [n]} \eta_i \left( \inner{v,h^i}^4 + \inner{v,\tl{h}^i}^4 \right),
$$
where the subtraction is replaced with addition.

Now, we partition the entries of vector $v = A^\top u \in \R^k$ into different vectors $v_l$ according to the magnitude of entries (this is very similar to Claim~\ref{claim2}). In particular, we partition entries (inner products) $v_j = \langle u,a_j \rangle, j \in [k]$, into $t+1$ buckets ($t:=\lceil \log_2 \sqrt{d} \rceil$) where (similar to Definition~\ref{def:Buckets})
\begin{align*}
K_0 &:= \left\{ j \in [k] : |\inner{u,a_j}| \leq \frac{1}{\sqrt{d}} \right\}, \\
K_l &:= \left\{ j \in [k] : |\inner{u,a_j}| \in \left(  \frac{2^{l-1}}{\sqrt{d}}, \frac{2^l}{\sqrt{d}} \right]  \right\}, \quad l \in [t].
\end{align*}
In addition, we merge the buckets $K_0$, $K_1,\dotsc,K_{\frac{1}{2} \log \log d}$ into $K_0$. This means $K_0$ now contains all $j$'s with inner product
$$|\inner{u,a_j}| \le \frac{\sqrt{\log d}}{\sqrt{d}},$$ and $K_l$'s for $1 \le l \le \frac{1}{2} \log\log d$ are empty. Now, let $v_l$ denote the restriction of vector $v$ to entries indexed by $K_l$, i.e.,
$$ v_l(j) :=
\left\{\begin{array}{ll}
v(j), & j \in K_l, \\
0, & j \notin K_l.
\end{array}\right.
$$
Let $p_l := 2^{l-1}$. By RIP property of matrix $A$, and exploiting Lemma~\ref{lem:bucketsRIP}, the number of nonzero entries in $v_l$ is bounded as
$$\|v_l\|_0 = |K_l| \leq O \left( \frac{d}{p_l^2} \right), \quad l > \frac{1}{2} \log\log d.$$

Exploiting the above partitioning, the term $\inner{v,h^i}^4$ in summation $Q$ can be upper bounded as
$$
\inner{v,h^i}^4 = \Bigl( \sum_{l=0}^t \inner{v_l,h^i} \Bigr)^4
\leq \Bigl( (t+1) \sum_{l=0}^t \inner{v_l,h^i}^2 \Bigr)^2
\leq (t+1)^3 \sum_{l=0}^t \inner{v_l,h^i}^4,
$$
where the equality is concluded from the fact that nonzero values of $v_l$'s are derived from partitioning of values of $v$, and Cauchy-Schwartz inequality is exploited in the last two steps. Applying this upper bound on $Q$, we would like to bound
$$
Q' := \frac{1}{n} \sum_{i\in [n]} \eta_i (t+1)^3 \sum_{l=0}^t \left( \inner{v_l,h^i}^4 + \inner{v_l,\tl{h}^i}^4 \right).
$$
In order to bound $Q'$, we break it into sum of $t+1$ terms as $Q' = \sum_{l=0}^t Q'_l$ where
$$
Q'_l := \frac{1}{n} (t+1)^3 \sum_{i \in [n]} \eta_i  \left( \inner{v_l,h^i}^4+\inner{v_l,\tl{h}^i}^4 \right).
$$

All terms $Q'_l$ can be bounded in the same way as Claim~\ref{claim:gaussian4th}. Especially, directly from Claim~\ref{claim:gaussian4th}, we have
$$Q'_0 \le \tilde{O} \left( \frac{s^2}{n}+\sqrt{\frac{s^4}{d^3n}} \right).$$
For the other terms $Q'_l, l > \frac{1}{2} \log \log d$, we need to analyze the tail behavior of $\inner{v_l,h^i}^4$. The tail behavior of this variable is affected by two phenomena: 1) the size of intersection of the supports of $v_l$ and $h^i$, and 2) given the intersection, the tail behavior of
\begin{equation} \label{eqn:SparseICA proof2}
\inner{v_l,h^i} = \sum_{j \in [k]: s^i[j]=1} v_l[j] g^i[j],
\end{equation}
which is a sum of subgaussian random variables.
Recall that $h^i[j] = s^i[j]g^i[j]$ where $s^i \in \R^k$ with i.i.d. Bernoulli random entries specifies the support of $h^i$.

The first part (the intersection of supports) can be bounded by Chernoff bound as
$$\Pr \Bigl[ \sum_{j \in [k]} s^i[j] \ge (1+\delta)s \Bigr] \le \left( \frac{e^\delta}{(1+\delta)^{(1+\delta)}} \right)^s.$$
The second part follows from subgaussian concentrations bounds. Let $\theta_l := \frac{2^l}{\sqrt{d}}$.
 For bucket $K_l$, and subsequently $Q'_l$ where $v_l$ has entries in the interval $(\theta_l/2,\theta_l]$, we discuss the tail behavior in two cases where $1/\theta_l^2 \ge s$ and $1/\theta_l^2\le s$.

{\bf Case 1} ($1/\theta_l^2 \ge s$):
In this case, most of $\inner{v_l,h^i}^4$ are of size $s^2/k^2$ which is very small.
For any $q \in \bigl[ \sqrt{s/(k\theta_l^2)} \polylog(n), s \bigr]$, since the summation in~\eqref{eqn:SparseICA proof2} is $\sqrt{s} \theta_l$-subgaussian, with probability at least $1 - e^{-\tilde{\Omega}(q)}$, we have
$$\inner{v_l,h^i}^4 \in \left(q^4\theta_l^4/2, q^4\theta_l^4 \right].$$
Therefore in this range, with probability at least $1 - e^{-\tilde{\Omega}(1/\theta_l^2)}$, the summation $Q'_l$ is bounded by
$$\frac{1}{n}\tilde{O} \left( \frac{q^4\theta_l^4}{\theta_l^2 q} \right) = \tilde{O} \left( \frac{q^3 \theta_l^2}{n} \right) \le \tilde{O} \left( \frac{s^2}{n} \right),$$
where the last inequality uses the fact that $\theta_l^2\le 1/s$.

For any $q \in \bigl( s, \sqrt{s/\theta_l^2}\log^2 n \bigr]$, since the summation in~\eqref{eqn:SparseICA proof2} is $\sqrt{s} \theta_l$-subgaussian, with probability at least $1 - e^{-\tilde{\Omega}(q^2/s)}$, we have
$$\inner{v_l,h^i}^4 \in \left(q^4\theta_l^4/2, q^4\theta_l^4 \right].$$
Therefore in this range, with probability at least $1 - e^{-\tilde{\Omega}(1/\theta_l^2)}$, the summation $Q'_l$ is bounded by
$$\frac{1}{n}\tilde{O} \left( q^4\theta_l^4 \frac{1}{\theta_l^2 q^2/s} \right) = \tilde{O} \left( \frac{q^2 \theta_l^2 s}{n} \right) \le \tilde{O} \left( \frac{s^2}{n} \right),$$
where the last inequality uses the fact that $q^2 = \tilde{O}(s/\theta_l^2)$.

When $q > \sqrt{s/\theta_l^2}\log^2 n$, there are no term $\inner{v_l,h^i}^4$ in this range  with high probability. Therefore, in the first case, by doing union bound $Q'_l$ is always bounded by
$$\tilde{O} \left( \frac{s^2}{n} \right)+o \left( \frac{s^4}{d^3 n} \right).$$

{\bf Case 2} ($1/\theta_l^2 \le s$):
In this case, again most of $\inner{v_l,h^i}^4$ are of size $s^2/k^2$ which is very small.
The only difference with case 1 is the two ranges where instead of being separated at $s$, they are separated at $1/\theta_l^2$ because there are at most $\tilde{O}(1/\theta_l^2)$ nonzero entries in $v_l$ as shown earlier.

For any $q \in \bigl[ \sqrt{s/(k\theta_l^2)} \polylog(n), 1/\theta_l^2 \bigr]$, since the summation in~\eqref{eqn:SparseICA proof2} is $\sqrt{s} \theta_l$-subgaussian, with probability at least $1 - e^{-\tilde{\Omega}(q)}$, we have
$$\inner{v_l,h^i}^4 \in \left(q^4\theta_l^4/2, q^4\theta_l^4 \right].$$
Therefore in this range, with probability at least $1 - e^{-\tilde{\Omega}(1/\theta_l^2)}$, the summation $Q'_l$ is bounded by
$$\frac{1}{n}\tilde{O} \left( \frac{q^4\theta_l^4}{\theta_l^2 q} \right) = \tilde{O} \left( \frac{q^3 \theta_l^2}{n} \right) \le \tilde{O} \left( \frac{s^2}{n} \right),$$
where the last inequality uses the fact that $\theta_l^2\le 1/s$.

For any $q \in \bigl( 1/\theta_l^2, \sqrt{s/\theta_l^2}\log^2 n \bigr]$, since the summation in~\eqref{eqn:SparseICA proof2} is $\sqrt{s} \theta_l$-subgaussian, with probability at least $1 - e^{-\tilde{\Omega}(q^2\theta_l^2)}$, we have
$$\inner{v_l,h^i}^4 \in \left(q^4\theta_l^4/2, q^4\theta_l^4 \right].$$
Therefore in this range, with probability at least $1 - e^{-\tilde{\Omega}(1/\theta_l^2)}$, the summation $Q'_l$ is bounded by
$$\frac{1}{n}\tilde{O} \left( q^4\theta_l^4 \frac{1}{\theta_l^2 q^2\theta_l^2} \right) = \tilde{O} \left( \frac{q^2}{n} \right) \le \tilde{O} \left( \frac{s^2}{n} \right),$$
where the last inequality uses the fact that $q^2 = \tilde{O}(s/\theta_l^2) \leq \tilde{O}(s^2)$.

When $q > \sqrt{s/\theta_l^2}\log^2 n$, there are no term $\inner{v_l,h^i}^4$ in this range  with high probability. Therefore, in the second case, by doing union bound $Q'_l$ is always bounded by
$$\tilde{O} \left( \frac{s^2}{n} \right)+o \left( \frac{s^4}{d^3 n} \right).$$

Combining the bounds on all terms finishes the proof.
\eprf

In the next claim we bound the perturbation of the 2nd order term $T$. Note that this is the sparse version of Claim~\ref{claim:ICA2nd}.

\begin{claim}
\label{claim:sparseICA2nd}
Consider the same sparse setting as in Theorem~\ref{lemma:sparseICA}.
Given $n$ samples $x^i = Ah^i, i \in [n],$ where $\|A\| \le O(\sqrt{k/d})$, for $T$ in~\eqref{eqn:ICA_2nd order term} and the empirical estimate $\widehat{T}$ in~\eqref{eqn:ICA_2nd order term empirical}, if $n\ge d$, we have with high probability
$$\|\widehat{T} - T\| \le\tilde{O} \left(\sqrt{\frac{s^4}{d^3n}} \right).$$
\end{claim}

\bprf
The proof is very similar to Claim~\ref{claim:ICA2nd}. Recall $W := \frac{1}{n}\sum_{i=1}^n x^i (x^i)^\top$.
We prove the result for the first term
$$\widehat{T}_1[i_1,i_2,i_3,i_4] = W_{i_1,i_2}W_{i_3,i_4},$$
or equivalently $\widehat{T}_1 = W\otimes W$. The analysis for the other two terms follow similarly from symmetry.
As in~\eqref{eqn:outerproduct}, we have
$$
\|\widehat{T}_1-T_1\| \le 2\|W-\E[W]\| \cdot \|\E[W]\|+\|W-\E[W]\|^2.
$$

We bound $\|W-\E[W]\|$ by Matrix Bernstein's inequality. As in Claim~\ref{claim:ICA2nd}, we first construct $$W' = \frac{1}{n}\sum_{i=1}^n x^i(x^i)^\top \mathbf{1}_{\|x^i\| \le O(\sqrt{s}\log n)},$$
 where $\mathbf{1}_{\|x^i\| \le O(\sqrt{s}\log n)}$ is an indicator variable. Since $x = Ah$ and entries of $h$ are subgaussian, the indicator variables are 1 with probability $1-n^{-\log n}$.
Therefore $W$ and $W'$ are equal with high probability at it suffices to apply Matrix Bernstein's bound on $W'$.

For the summation $W'$, the norm of each term is bounded by $\tilde{O}(s/n)$, and for the variance term, we have
$$\E \left[ W'(W')^\top \right] = \frac{1}{n}\E \left[ \|x^i\|^2x^i(x^i)^\top\mathbf{1}_{\|x^i\| \le O(\sqrt{s}\log n)} \right] \preceq \frac{1}{n}\tilde{O}(s)\E \left[ x^i(x^i)^\top \right] = \frac{1}{n}\tilde{O}(s^2/k)AA^\top.$$
Since $\|A\| \le O(\sqrt{k/d})$, it is concluded that the variance is bounded by $\tilde{O}(s^2/dn)$. Therefore, Matrix Bernstein's inequality implies that with probability at least $1-d/n$,
$$\|W'-\E[W']\| \le \tilde{O} \left( \frac{s}{n} + \frac{s}{\sqrt{dn}} \right).$$
Since $W$ is equal to $W'$ with high probability and $\|\E[W]-\E[W']\|$ is negligible, we also have $\|W-\E[W]\| \le \tilde{O}(s/\sqrt{dn})$ (when $n\ge d$).

On the other hand, $\E[W] = \frac{s}{k} AA^\top$, and therefore, $\|\E[W]\| \leq s/d$. From \eqref{eqn:outerproduct}, we have
$$\|\widehat{T}_1-T_1\| \le\tilde{O} \left(\sqrt{\frac{s^4}{d^3n}} \right).$$
\eprf

\end{document}